\PassOptionsToPackage{unicode}{hyperref}
\PassOptionsToPackage{hyphens}{url}

\documentclass[11pt]{article}
\usepackage{xcolor}
\usepackage{amsmath,amssymb}

\usepackage{iftex}
\ifPDFTeX
  \usepackage[T1]{fontenc}
  \usepackage[utf8]{inputenc}
  \usepackage{textcomp} 
\else 
  \usepackage{unicode-math} 
  \defaultfontfeatures{Scale=MatchLowercase}
  \defaultfontfeatures[\rmfamily]{Ligatures=TeX,Scale=1}
\fi
\usepackage{lmodern}
\ifPDFTeX\else
\fi
\IfFileExists{upquote.sty}{\usepackage{upquote}}{}
\IfFileExists{microtype.sty}{
  \usepackage[]{microtype}
  \UseMicrotypeSet[protrusion]{basicmath} 
}{}
\makeatletter
\@ifundefined{KOMAClassName}{
  \IfFileExists{parskip.sty}{%
    \usepackage{parskip}
  }{
    \setlength{\parindent}{0pt}
    \setlength{\parskip}{6pt plus 2pt minus 1pt}}
}{
  \KOMAoptions{parskip=half}}
\makeatother
\usepackage{longtable,booktabs,array}
\usepackage{calc} 
\usepackage{etoolbox}
\makeatletter
\patchcmd\longtable{\par}{\if@noskipsec\mbox{}\fi\par}{}{}
\makeatother
\IfFileExists{footnotehyper.sty}{\usepackage{footnotehyper}}{\usepackage{footnote}}
\makesavenoteenv{longtable}
\usepackage{graphicx}
\usepackage{caption}
\makeatletter
\newsavebox\pandoc@box
\newcommand*\pandocbounded[1]{
  \sbox\pandoc@box{#1}%
  \Gscale@div\@tempa{\textheight}{\dimexpr\ht\pandoc@box+\dp\pandoc@box\relax}%
  \Gscale@div\@tempb{\linewidth}{\wd\pandoc@box}%
  \ifdim\@tempb\p@<\@tempa\p@\let\@tempa\@tempb\fi
  \ifdim\@tempa\p@<\p@\scalebox{\@tempa}{\usebox\pandoc@box}%
  \else\usebox{\pandoc@box}%
  \fi%
}
\def\fps@figure{htbp}
\makeatother
\usepackage{bookmark}
\IfFileExists{xurl.sty}{\usepackage{xurl}}{} 
\hypersetup{
  hidelinks,
  pdfcreator={LaTeX via pandoc}}

\author{}
\date{}

\usepackage[left=2.5cm,right=2.5cm,top=2.5cm,bottom=2.5cm]{geometry}

\usepackage{authblk}

\begin{document}


\title{\parbox{0.9\textwidth}{\centering\fontsize{20pt}{24pt}\selectfont A new dual-population constrained multi-objective evolutionary optimization algorithm with repair constraint handling for structural optimization}}

\author[1]{Fardad Homafar}
\author[1,2,*]{Jasmin Jelovica}

\affil[1]{\fontsize{10pt}{12pt}\selectfont Department of Mechanical Engineering, The
University of British Columbia, 6250 Applied Science Ln, Vancouver, V6T
1Z4, Canada}
\affil[2]{Department of Civil Engineering, The
University of British Columbia, 6250 Applied Science Ln, Vancouver, V6T
1Z4, Canada}

\maketitle
\begingroup
\renewcommand{\thefootnote}{\fnsymbol{footnote}}
\footnotetext[1]{Corresponding author.\\
\textit{Email address:}
\href{mailto:jasmin.jelovica@ubc.ca}{\texttt{jasmin.jelovica@ubc.ca}}
(Jasmin Jelovica)}
\endgroup





\noindent\rule{\textwidth}{0.4pt}

\textbf{Abstract:}

Structural optimization problems often involve a large number of
decision variables and highly non-convex feasible regions, making
convergence to the true Pareto front extremely challenging. Even when
convergence is achievable, it typically requires thousands of function
evaluations, resulting in significant computational cost. This
highlights the need for efficient and robust optimization algorithms for
real-world engineering applications. In this study, we introduce a novel
constrained multi-objective evolutionary algorithm, termed DPCME. The
algorithm employs two interacting populations that exchange information,
enabling effective global exploration and reducing the risk of
convergence to local optima. To further enhance performance, a recent
repair-based constraint-handling technique is incorporated, and
alternative repair approaches are proposed and systematically evaluated.
The proposed algorithm is tested on three engineering problems: the
72-bar truss, the 120-bar truss, and a chemical tanker structure, each
involving hundreds of nonlinear failure constraints. Its performance is
evaluated against state-of-the-art constrained multi-objective
optimization algorithms from the latest PlatEMO package. A total of 43
algorithms are initially tested, from which the 12 best-performing
methods are selected for detailed comparison. The results demonstrate
that DPCME achieves superior or competitive convergence and diversity
across all test cases, and that the inclusion of repair-based constraint
handling further improves its performance.

\noindent\rule{\textwidth}{0.4pt}


\section{Introduction}\label{introduction}

Identifying global optima in engineering design is crucial for
maximizing product performance. However, real-world structural
engineering optimization problems are often highly complex due to their
nonconvex functions, conflicting objectives, and discontinuous design
domains. In such challenging scenarios, metaheuristic optimization
algorithms are typically the methods of choice. Over the past few years,
numerous metaheuristic algorithms have been developed, inspired by
various natural, physical, and biological processes. Many of these
approaches emulate the behaviors and cooperative strategies of animals,
for instance, the Coati Optimization Algorithm (COA)[\textcolor{blue}{\hyperlink{ref:1}{1}}] , Fire Hawk
Optimizer (FHO) [\textcolor{blue}{\hyperlink{ref:2}{2}}], Mountain Gazelle Optimizer (MGO) [\textcolor{blue}{\hyperlink{ref:3}{3}}],
Gannet Optimization Algorithm (GOA) [\textcolor{blue}{\hyperlink{ref:4}{4}}], Aquila Optimizer (AO)
[\textcolor{blue}{\hyperlink{ref:5}{5}}], and Honey Badger Algorithm (HBA) [\textcolor{blue}{\hyperlink{ref:6}{6}}]. Other recent
algorithms draw inspiration from physical phenomena, including the
Thermal Exchange Optimization (TEO) [\textcolor{blue}{\hyperlink{ref:7}{7}}], Crystal Structure Algorithm
(CryStAl) [\textcolor{blue}{\hyperlink{ref:8}{8}}], Chaos Game Optimization (CGO) [\textcolor{blue}{\hyperlink{ref:9}{9}}], Atomic Orbital
Search (AOS) [\textcolor{blue}{\hyperlink{ref:10}{10}}], Henry Gas Solubility Optimization (HGSO)
[\textcolor{blue}{\hyperlink{ref:11}{11}}], War Strategy Optimization (WSO) [\textcolor{blue}{\hyperlink{ref:12}{12}}], and Arithmetic
Optimization Algorithm (AOA) [\textcolor{blue}{\hyperlink{ref:13}{13}}].

Another important development is surrogate-based global optimization,
which relies on constructing computationally inexpensive approximations
of expensive objective and constraint functions using models such as
polynomial response surfaces, radial basis function (RBF) networks,
Kriging or Gaussian process (GP) models, and neural networks [\textcolor{blue}{\hyperlink{ref:14}{14}}].
These surrogates are typically embedded within optimization frameworks
such as Efficient Global Optimization (EGO) [\textcolor{blue}{\hyperlink{ref:15}{15}}], RBF-based
trust-region methods [\textcolor{blue}{\hyperlink{ref:16}{16}}, \textcolor{blue}{\hyperlink{ref:17}{17}}], and multi-objective extensions
including ParEGO [\textcolor{blue}{\hyperlink{ref:18}{18}}], or integrated into surrogate-assisted
evolutionary algorithms [\textcolor{blue}{\hyperlink{ref:19}{19}}]. Surrogate models have also been
combined with nature-inspired metaheuristics to reduce evaluation cost
while preserving global search capability [\textcolor{blue}{\hyperlink{ref:20}{20}}]. Nevertheless, in
many engineering fields, including structural engineering,
computationally intensive full-order models remain the most accurate and
reliable analysis tools. At the same time, metaheuristic optimization
algorithms require a large number of function evaluations to achieve
satisfactory convergence. When the analysis is computationally
demanding, the resulting cost often allows only a limited number of
evaluations, making rapid progress through the objective space
essential. Furthermore, stringent design requirements and practical time
constraints may necessitate the premature termination of optimization
runs.

Among various metaheuristic frameworks, \emph{evolutionary algorithms
(EAs)} represent one of the most prominent and well-established classes.
These algorithms are inspired by the principles of natural evolution and
operate on a population of candidate solutions that evolve over
successive generations. Through mechanisms such as selection, crossover,
and mutation, EAs effectively balance exploration and exploitation in
the search space. Well-known representatives of this family include the
Genetic Algorithm (GA) [\textcolor{blue}{\hyperlink{ref:21}{21}}, \textcolor{blue}{\hyperlink{ref:22}{22}}], Differential Evolution (DE)
[\textcolor{blue}{\hyperlink{ref:23}{23}}], Evolution Strategy (ES) [\textcolor{blue}{\hyperlink{ref:24}{24}}], and Covariance Matrix
Adaptation Evolution Strategy (CMA-ES) [\textcolor{blue}{\hyperlink{ref:25}{25}}]. Owing to their
population-based structure and adaptability, evolutionary algorithms
have been widely applied to complex, nonlinear, and multi-objective
optimization problems across diverse engineering domains.
Multi-population evolutionary algorithms (MPEAs) exploit multiple
coexisting subpopulations (or ``islands'') to concurrently explore
different regions of the search space. A well-known example is Multiple
Populations for Multiple Objectives (MPMO), where each subpopulation is
dedicated to a single objective, yet they exchange information via a
shared archive of elite solutions to approximate the Pareto front more
effectively. Periodic migration or knowledge sharing ensures that
promising individuals from one population can influence others, thus
facilitating cross-objective learning and improving global convergence
[\textcolor{blue}{\hyperlink{ref:26}{26}}]. Beyond MPMO, many multipopulation methods adopt
co-evolutionary strategies or reference exchange to enable richer
knowledge transfer. For example, Multipopulation Coevolutionary Strategy
for Multiobjective Immune Algorithm (often abbreviated as CONNIA or
similar) maintains two or more evolving populations that
intercommunicate via cooperation operators: the weaker or less diverse
subpopulation receives ``reference'' or ``experience'' solutions from
the stronger one to guide its search [\textcolor{blue}{\hyperlink{ref:27}{27}}]. Another variant,
Multi-Population Coevolution Immune Algorithm (MCIA), uses three
interacting populations (e.g. B-cell, T-cell, helper populations) and
shares elite individuals, crossover/mutation operators, or fitness cues
across them to help all populations avoid stagnation and improve
convergence and diversity [\textcolor{blue}{\hyperlink{ref:28}{28}}]. Building upon this concept, the
Improved Multi-Tasking Constrained Multi-objective Optimization (IMTCMO)
algorithm maintains two interacting populations, each evolving
simultaneously. Knowledge transfer occurs by exchanging promising
solutions between the populations, while each population applies
selection mechanisms tailored to handle constraints effectively. This
multitasking approach allows the algorithm to explore feasible and
infeasible regions concurrently, enhancing diversity and convergence in
constrained multi-objective problems [\textcolor{blue}{\hyperlink{ref:29}{29}}]. Similarly, the
Constrained Multi-objective Evolutionary algorithm with Global and Local
Auxiliary task (CMEGL) algorithm introduces a multitasking strategy by
incorporating a main task and two auxiliary tasks. The main task
addresses the constrained multi-objective optimization problem, while
the global auxiliary task explores the entire search space to identify
promising feasible regions, and the local auxiliary task enhances local
diversity by utilizing promising infeasible solutions. Knowledge
transfer occurs among these tasks to guide the search process, improving
the algorithm\textquotesingle s performance in handling complex feasible
regions [\textcolor{blue}{\hyperlink{ref:30}{30}}]. Nonetheless, many recently proposed algorithms have
not been thoroughly validated on realistic, complex structural
engineering problems with stringent physical failure constraints. Even
when tested on commonly used engineering benchmark problems, the
resulting Pareto fronts are often suboptimal, indicating substantial
room for further improvement.

In the presence of constraints, metaheuristic algorithms typically rely
on constraint-handling techniques (CHTs) to effectively explore the
feasible region and identify constrained optima [\textcolor{blue}{\hyperlink{ref:31}{31}}--\textcolor{blue}{\hyperlink{ref:34}{34}}]. The most
common CHTs are penalty functions, including static, dynamic, and
adaptive variants. Repair-based CHTs can be more efficient; however,
they are less commonly used because they often rely on problem-specific
knowledge. For example, in stacking-sequence optimization of composite
laminates [\textcolor{blue}{\hyperlink{ref:35}{35}}], constraints limit the maximum number of consecutive
plies with identical fiber orientations. When this requirement is
violated, the solution is modified using a set of predefined heuristic
rules until feasibility is restored. This process typically requires
substantial domain expertise, which represents a major limitation of
repair-based approaches and restricts their general applicability. To
address this issue, a more recent repair method was introduced in which
high-performing infeasible solutions are repaired using information from
other population members that do not violate the same constraints
[\textcolor{blue}{\hyperlink{ref:36}{36}}]. This method operates in a semi-autonomous manner, as it
leverages population knowledge to modify valuable infeasible solutions
rather than relying on user-defined heuristics. The user is only
required to specify which variables affect which constraints---a
limitation that was recently overcome through a data-driven approach
that enables automatic construction of this mapping [\textcolor{blue}{\hyperlink{ref:37}{37}}]. The
approach was also adapted for the MOEA/D algorithm [\textcolor{blue}{\hyperlink{ref:38}{38}}]. However,
alternative repair approaches exist that have not yet been explored and
may lead to further performance improvements. In particular, it remains
unclear which population subgroups should be used to repair promising
infeasible solutions and how these solutions should be prioritized---two
issues that warrant systematic investigation.

In this study, we propose a new constrained multi-objective optimization
algorithm that is tested on structural engineering problems. The
proposed approach employs two parallel populations, referred to as the
main and auxiliary populations, that exchange knowledge throughout the
evolutionary process. This interaction enables the algorithm to
effectively identify high-quality feasible solutions while reducing the
likelihood of premature convergence to local optima. In addition, we
further enhance the algorithm's performance through a few new autonomous
repair approaches for constraint handling. To evaluate the effectiveness
of the proposed approach, we conduct experiments on three test problems:
two well-known truss optimization problems and a more challenging
structural optimization of a ship chemical tanker. These problems
involve between 144 and 376 nonlinear structural failure constraints.
All of these are bi-objective optimization problems; however, the
algorithm is readily applicable to problems with a higher number of
objectives. The performance of the proposed algorithm is compared
against the best-performing constrained multi-objective optimization
algorithms available in the PlatEMO platform [\textcolor{blue}{\hyperlink{ref:39}{39}}].

\section{Methodology}\label{methodology}

\subsection{General framework}\label{general-framework}

A constrained multi-objective optimization problem (CMOP) can be
formulated as follows:

\begin{equation}
\begin{aligned}
\operatorname{minimize}\quad
  &F\left(\mathbf{x}\right)
   = \left(f_{1}\left(\mathbf{x}\right),\ldots,
            f_{m}\left(\mathbf{x}\right)\right)^{T}, \\
\text{subject to}\quad
  &g_{i}\left(\mathbf{x}\right) \geq 0,
   \quad i=1,\ldots,q, \\
  &h_{j}\left(\mathbf{x}\right) = 0,
   \quad j=1,\ldots,p, \\
  &\mathbf{x} \in \mathbb{R}^{n}.
\end{aligned}
\tag{1}
\end{equation}

where \emph{F(x)} is an \emph{m}-dimensional objective vector, while
\(g_{i}\left( \mathbf{x} \right)\) and
\(h_{j}\left( \mathbf{x} \right)\) represent the inequality and equality
constraints, respectively. \(\mathbf{x} \in \mathbb{R}^{n}\) denotes an
\emph{n}-dimensional decision vector.

The feasible region \(\mathrm{\Omega}\) is defined as the set
\(\{\mathbf{x}|g_{i}\left( \mathbf{x} \right) \geq 0,\ i = 1,\ldots,q\)
and \(h_{j}\left( \mathbf{x} \right) = 0,\ j = 1,\ldots,p\}\), while its
complement is denoted as the infeasible region
\(\widehat{\mathrm{\Omega}}\). For any two feasible solutions
\(\mathbf{x}^{1},\mathbf{x}^{2} \in \mathrm{\Omega}\),
\(\mathbf{x}^{1}\) is said to dominate \(x^{2}\) if
\(f_{k}\left( \mathbf{x}^{1} \right) \leq f_{k}\left( \mathbf{x}^{2} \right)\)
for all \(k \in \left\{ 1,\ldots,m \right\}\), and the inequality is
strict for at least one objective \emph{l}. A solution
\(\mathbf{x}^{*} \in \mathrm{\Omega}\) is Pareto optimal if no other
feasible solution dominates it. The set of all Pareto optimal solutions
constitutes the Pareto set (PS), and its mapping in the objective space
forms the Pareto front (PF), defined as
\(PF = \left\{ F\left( \mathbf{x} \right)|\mathbf{x} \in PS \right\}\).
In practice, multi-objective evolutionary algorithms (MOEAs) aim to
approximate the PF by iteratively evolving a population of candidate
solutions across generations.

\subsection{Dual-population constrained multi-objective evolutionary
algorithm
(DPCME)}\label{dual-population-constrained-multi-objective-evolutionary-algorithm-dpcme}

We propose here dual-population constrained multi-objective evolutionary
(DPCME) algorithm, which operates using two co-evolving populations that
interact with each other for the purpose of distinct evolutionary goals.
The main population focuses on the original multi-objective optimization
problem and uses NSGA-II--style [\textcolor{blue}{\hyperlink{ref:40}{40}}] selection and a more
exploitative operator to promote convergence and diversity along the
Pareto front. In contrast, the auxiliary population is designed purely
for exploration: it uses an inverse-crowding fitness measure, standard
genetic operators, and a constraint-relaxing selection mechanism
governed by a dynamic parameter. This allows the auxiliary task to
search loosely constrained or under-explored regions that the main
population would normally discard. Offspring exchange between the two
populations enables the exploratory behavior of the auxiliary task to
guide the main search while still preserving strong convergence
pressure. Both populations use the same population size, which remains
constant throughout the optimization process. The population size for
each population is set to 100, a value commonly adopted in the
literature.

Literature has shown that adaptive threshold as CHT performs better than
various penalty functions for engineering problems [\textcolor{blue}{\hyperlink{ref:37}{37}}, \textcolor{blue}{\hyperlink{ref:41}{41}}], thus we
use a dynamic constraint threshold parameter, \(\beta_{t}\) as the
baseline CHT for DPCME. It is calculated as the mean constraint
violation in the main population and is later used to determine the
feasibility of individuals in the auxiliary task. Section 2.3 presents
the repair-based CHT used in DPCME, which is applied to a designated
portion of the population in addition to the adaptive threshold
mechanism already embedded in the algorithm for the rest of the
population.

During the iterative optimization process, offspring are generated
separately for each task. In the main task, individuals are selected
using binary tournament selection, guided by non-dominance and crowding
distance, and then recombined using genetic operators (simulated binary
crossover and polynomial mutation) to produce new solutions. For the
auxiliary task, the selection process emphasizes exploration by relying
solely on crowding distance. The same genetic operators are applied to
generate auxiliary offspring.

Once offspring are evaluated, the algorithm proceeds to evolve both
populations through environmental selection. To avoid unwanted knowledge
transfer, environmental selection for the main task is executed in two
stages. Inspired by CMEGL algorithm [\textcolor{blue}{\hyperlink{ref:30}{30}}], environmental selection is
applied to the combined set of main population and auxiliary offspring,
and subsequently to the union of the main population and its own
offspring. This staged approach enables the main task to benefit from
auxiliary diversity without being overwhelmed by it. Meanwhile, the
auxiliary task undergoes its own environmental selection over a combined
pool of its current population, its offspring, and the offspring from
the main task.

Constraint handling in both tasks is guided by feasibility-based rules.
In the main task, feasible solutions are always prioritized over
infeasible ones. Each individual\textquotesingle s fitness is computed
using a hybrid metric that combines a dominance-based rank, following
the approach used in SPEA2 [\textcolor{blue}{\hyperlink{ref:42}{42}}], and a density estimation term
calculated from the inverse distance to the \emph{k}-th nearest neighbor
in objective space. This promotes both convergence and diversity. When
feasible solutions are insufficient to fill the desired population size
\emph{N}, the best infeasible individuals (with the lowest fitness
values) are included. If there are enough feasible individuals,
survivors are selected based on crowding distance to preserve diversity.

In the auxiliary task, feasibility is determined relative to the dynamic
constraint threshold \(\beta_{t}\); individuals whose constraint
violations fall below this threshold are deemed feasible. Non-dominated
sorting is applied to the feasible solutions, and those assigned to any
front other than the worst one are designated as potential survivors.
This ensures that poorly performing feasible solutions are excluded from
consideration. If the number of potential survivors exceeds \emph{N}, a
second round of non-dominated sorting is applied to this subset, and
survivors are selected based on rank and, if needed, crowding distance
as a tie-breaker. Non-dominated sorting is applied several times
throughout the algorithm due to its proven effectiveness in achieving
high-quality solutions in engineering optimization [\textcolor{blue}{\hyperlink{ref:40}{40}}, \textcolor{blue}{\hyperlink{ref:43}{43}}-\textcolor{blue}{\hyperlink{ref:45}{45}}]. On
the other hand, if fewer than \emph{N} potential survivors are
available, all feasible individuals are first ranked based on
non-dominated sorting and crowding distance, and the top \emph{N} are
retained as the next generation of the auxiliary population. Just in
case there are fewer than \emph{N} feasible individuals, all of them are
returned as survivors. This flexible selection mechanism allows the
auxiliary task to maintain diversity while focusing on promising regions
of the search space under relaxed constraint handling.

{\def\LTcaptype{none} 
\begin{longtable}[]{@{}
  >{\raggedright\arraybackslash}p{(\linewidth - 0\tabcolsep) * \real{1.0000}}@{}}
\toprule\noalign{}
\begin{minipage}[b]{\linewidth}\raggedright
\textbf{Algorithm 1: Main algorithm}
\end{minipage} \\
\midrule\noalign{}
\endhead
\bottomrule\noalign{}
\endlastfoot
\begin{minipage}[t]{\linewidth}\raggedright
\textbf{Input:} Population size (\emph{N}), Max \# of Function
Evaluations (MaxFE)

\begin{enumerate}
\def\labelenumi{\arabic{enumi}.}
\item
  \(P_{1}\)← Generate \emph{N} random individuals
\item
  \(P_{2}\) ← Generate \emph{N} random individuals
\item
  Evaluate two populations
\item
  \(\beta_{t}\) ← mean(CV(infeasible individuals in \(P_{1}\)))
\item
  \(t = 0\)
\item
  \textbf{while} FE\textless MaxFE:
\item
  t ← t+1
\item
  \(O_{1}\ \)← Use the parents from \(P_{1}\) to generate \emph{N}/2
  offspring NSGAII style
\item
  Crowding ← Calculate crowding distance between the individuals in
  \(P_{2}\)
\item
  Mating Pool ← Tournament selection based on Crowding
\item
  \(O_{2}\) ← GA operator on the Mating Pool
\item
  Evaluate \(O_{1}\) and \(O_{2}\)
\item
  \(P_{1}\) ← EnvironmentalSelection\((O_{2} \cup P_{1})\)
\item
  \(P_{1}\) ← EnvironmentalSelection\((O_{1} \cup P_{1})\)
\item
  \(P_{2}\) ← EnvironmentalSelectionAT \((P_{2} \cup O_{1} \cup O_{2})\)
\item
  \(\beta_{t}\) ← mean(CV(infeasible individuals in \(P_{1}\)))
\item
  \textbf{end}
\end{enumerate}

\textbf{Output}: Feasible Pareto optimal solutions
\end{minipage} \\
\end{longtable}
}

{\def\LTcaptype{none} 
\begin{longtable}[]{@{}
  >{\raggedright\arraybackslash}p{(\linewidth - 0\tabcolsep) * \real{1.0000}}@{}}
\toprule\noalign{}
\begin{minipage}[b]{\linewidth}\raggedright
\textbf{Algorithm 2: Environmental selection}
\end{minipage} \\
\midrule\noalign{}
\endhead
\bottomrule\noalign{}
\endlastfoot
\begin{minipage}[t]{\linewidth}\raggedright
\textbf{Input:} Population, \# of survivors (\emph{NS})

\begin{enumerate}
\def\labelenumi{\arabic{enumi}.}
\item
  Fitness ← Calculate fitness value of the population.
\item
  Next ← Find feasible individuals indices of the population
\item
  \textbf{if} size(Next)\textless{}\emph{NS}:
\item
  sort Population by Fitness
\item
  outputPop ← Select NS best individuals
\item
  \textbf{else if} size(Next)\textgreater{}\emph{NS}:
\item
  Crowding ← Calculate crowding distance of fPopulation
\item
  Sort feasible individuals by Crowding
\item
  outputPop ← Select NS best individuals
\item
  \textbf{end}
\end{enumerate}

\textbf{Output:} Survived population (outputPop), Fitness
\end{minipage} \\
\end{longtable}
}

{\def\LTcaptype{none} 
\begin{longtable}[]{@{}
  >{\raggedright\arraybackslash}p{(\linewidth - 0\tabcolsep) * \real{1.0000}}@{}}
\toprule\noalign{}
\begin{minipage}[b]{\linewidth}\raggedright
\textbf{Algorithm 3: Environmental selection for auxiliary task}
\end{minipage} \\
\midrule\noalign{}
\endhead
\bottomrule\noalign{}
\endlastfoot
\begin{minipage}[t]{\linewidth}\raggedright
\textbf{Input:} Population, \# of survivors (\emph{NS}), Constraint
tolerance threshold (\(\beta_{t}\))

\begin{enumerate}
\def\labelenumi{\arabic{enumi}.}
\item
  fPopulation ← find individuals with \(CV \leq \beta_{t}\)
\item
  \textbf{if} fPopulation is empty:
\item
  outputPop ← {[}{]}
\item
  \textbf{end}
\item
  FrontNO,MaxFrontNO ← Apply non-dominated sorting on fPopulation
\item
  Selected ← Find individuals with FrontNo\textless MaxFrontNo
\item
  \textbf{if} size(selected)≥\emph{NS}:
\item
  Crowding ← Calculate crowding distance of selected individuals
\item
  Sort selected individuals by FrontNo. Use Crowding for tie break.
\end{enumerate}

\begin{enumerate}
\def\labelenumi{\arabic{enumi}.}
\item
  outputPop ← Select NS best individuals
\end{enumerate}

\begin{enumerate}
\def\labelenumi{\arabic{enumi}.}
\setcounter{enumi}{9}
\item
  \textbf{else:}
\item
  CrowDis ← Compute crowding distance for each individual in its own
  front NO.
\item
  rank ← Sort selected individuals by FrontNo. Use Crowding for tie
  break.
\item
  \textbf{if} size(rank)≥\emph{NS}:
\item
  outputPop ← Select best NS individuals from fPopulation based on the
  rank
\item
  \textbf{else:}
\item
  outputPop ← fPopulation
\item
  \textbf{end}
\item
  \textbf{end}
\end{enumerate}

\textbf{Output:} Survived population (outputPop)
\end{minipage} \\
\end{longtable}
}

\subsection{Modified adaptive repair approach for constraint
handling}\label{modified-adaptive-repair-approach-for-constraint-handling}

\begin{figure}
\centering
\makebox[\linewidth][c]{\includegraphics[width=2.57874in,height=2.30709in]{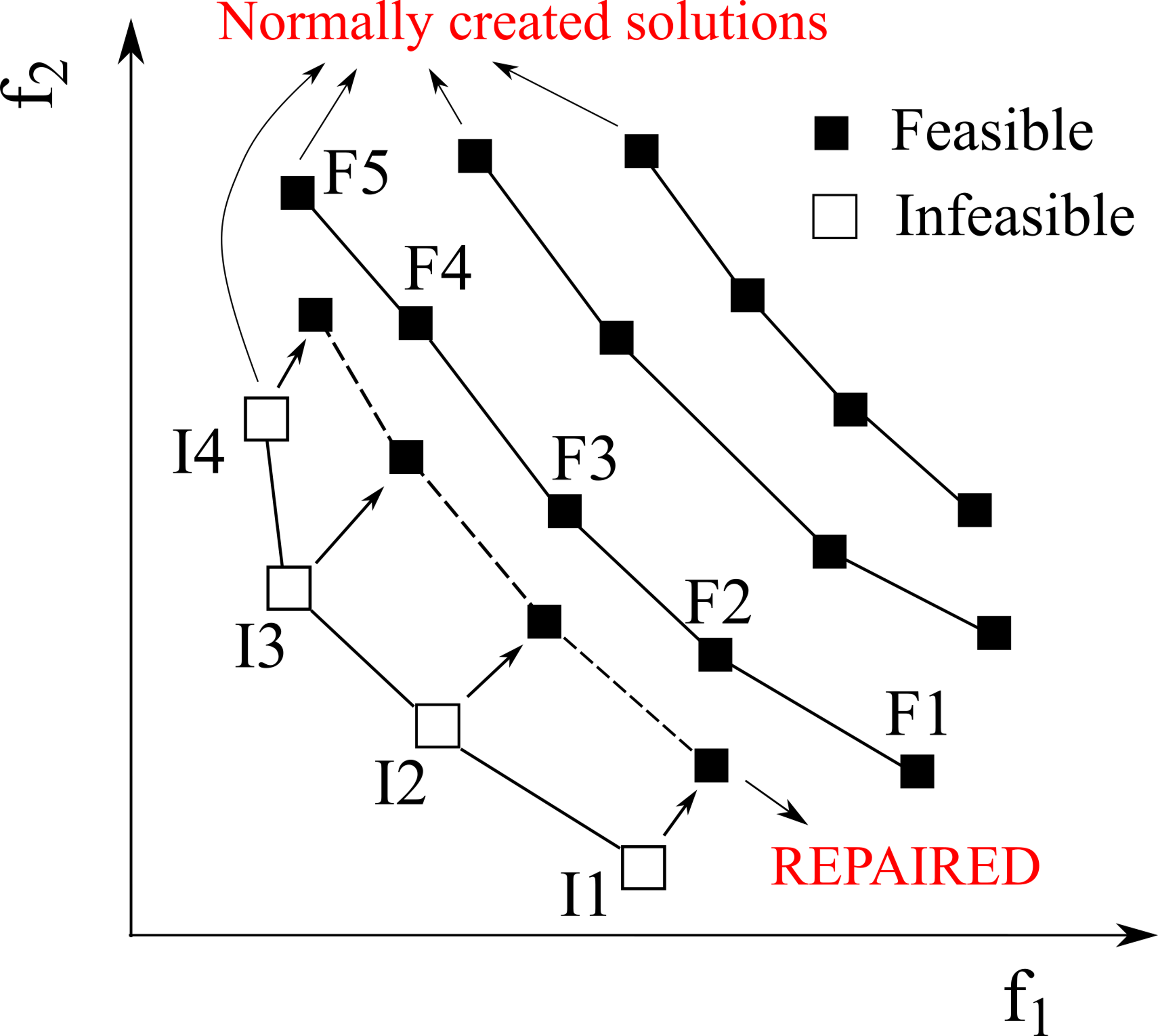}}
\caption{Aim of the repair CHT: Non-dominated infeasible
solutions after repair dominate ordinary feasible solutions.}
\label{fig:1}
\end{figure}

Another CHT implemented in the proposed DPCME algorithm is the
autonomous repair CHT [\textcolor{blue}{\hyperlink{ref:36}{36}}- \textcolor{blue}{\hyperlink{ref:38}{38}}], which replaces the values of
variables that cause constraint violations with corresponding values
from solutions in the population that do not violate the same
constraint. In this research, we extend the approach to explore
alternative repair approaches that have not been proposed previously,
yet they could hold promise for improving the performance. In each
generation, the repair algorithm receives two separate inputs: the
combined parent populations from both tasks and the combined offspring
populations from both tasks. Infeasible solutions being repaired are
called ``candidates'', while the ones used to repair them are called
``donors''. Candidates are selected from different groups of solutions
prioritizing non-domination rank and crowding distance. A desired repair
rate is specified, representing the percentage of solutions to be
repaired, decreasing the share of solutions created normally by the
algorithm. It is set to 10\% of the population size when at least one
feasible solution exists, and 50\% of the offspring size when no
feasible solution is available. These rates have been shown to lead to
best results in previous research [\textcolor{blue}{\hyperlink{ref:36}{36}}]. Prior to applying the repair
method, a mapping matrix between variables and constraints is specified.
The matrix identifies which decision variables most strongly influence
each constraint, indicating which variables should be adjusted when a
particular constraint is violated. The mapping matrix can be constructed
directly when explicit constraint functions are available. However, many
engineering problems do not provide such closed-form expressions. In
these cases, following the procedure outlined in [\textcolor{blue}{\hyperlink{ref:37}{37}}], the mapping
matrix can be derived using data-driven methods. As illustrated in
\textcolor{blue}{\hyperref[fig:1]{Fig. 1}}, the aim of the repair CHT is to transform originally infeasible
solutions that dominate the population in objective space into
non-dominated feasible solutions.

It can be assumed that using feasible donors to repair infeasible
candidates generally increases the likelihood of successful repair,
where ``success'' primarily means producing a feasible solution.
However, early in the optimization the entire population may be
infeasible, making feasible donors unavailable. In such situations,
infeasible solutions can be used for repair. This motivates the two main
repair scenarios: Repair during the infeasible phase (RI) and Repair
during the feasible phase (RF), summarized respectively in
\textcolor{blue}{\hyperref[tab:1]{Tables 1}} and \textcolor{blue}{\hyperref[tab:2]{2}}. RI1 and RF1 have been proposed previously [\textcolor{blue}{\hyperlink{ref:36}{36}}-\textcolor{blue}{\hyperlink{ref:38}{38}}], while the
others are new. To increase diversity and enlarge the pool of potential
candidates, the repair procedure is applied to a combined
parent-offspring population in RI3 and RI4. It is assumed that all
offspring are infeasible at RI phase − an assumption that is valid in
initial generations for highly constrained problems. In later stages of
the optimization, infeasible solutions typically violate fewer
constraints than in the initial phases and may even dominate feasible
solutions on the Pareto front. This motivates their use as donors in RF3
in addition to feasible solutions. A designated quota of candidates for
repair is obtained by selecting infeasible solutions using
non-domination sorting (NDS) as the primary criterion, with crowding
distance (CD) used as a tie-breaker. In some repair scenarios, the
Euclidean distance (ED) in the objective space replaces CD, because
closer donors are more likely to preserve the desirable location of the
candidate. As shown in \textcolor{blue}{\hyperref[fig:1]{Fig. 1}}, solution I1 is more likely to remain
close to its original position when repaired by F1 or F2, rather than by
F4 or F5. Similarly, F4 or F5 may be more suitable for repairing I4.
When the repaired solution remains near its original position and
becomes feasible, the optimization process could progress more
effectively. NDS is combined with ED in RI2 and RI3, while ED is used
without NDS in RI4 and RF2. In some scenarios only offspring population
is repaired, since this approach is simpler to implement. It is worth
noting that although the repair CHT increases the computational time of
the algorithm, this overhead is negligible, as the majority of the
computational cost in engineering applications is associated with
function evaluations. Moreover, if the proposed approach can reduce the
number of generations required to reach the same PF, the additional
runtime is well justified.

\begin{center}
\protect\phantomsection\label{tab:1}{}
Table 1. Repair strategies for fully infeasible populations (RI).
\end{center}

{\def\LTcaptype{none} 
\begin{longtable}[]{@{}
  >{\centering\arraybackslash}p{(\linewidth - 8\tabcolsep) * \real{0.1217}}
  >{\centering\arraybackslash}p{(\linewidth - 8\tabcolsep) * \real{0.2430}}
  >{\centering\arraybackslash}p{(\linewidth - 8\tabcolsep) * \real{0.2284}}
  >{\centering\arraybackslash}p{(\linewidth - 8\tabcolsep) * \real{0.1826}}
  >{\centering\arraybackslash}p{(\linewidth - 8\tabcolsep) * \real{0.2028}}@{}}
\toprule\noalign{}
\begin{minipage}[b]{\linewidth}\centering
Strategy
\end{minipage} & \begin{minipage}[b]{\linewidth}\centering
Candidates
\end{minipage} & \begin{minipage}[b]{\linewidth}\centering
Selection of candidates
\end{minipage} & \begin{minipage}[b]{\linewidth}\centering
Donors
\end{minipage} & \begin{minipage}[b]{\linewidth}\centering
Selection of donors
\end{minipage} \\
\midrule\noalign{}
\endhead
\bottomrule\noalign{}
\endlastfoot
RI1 [\textcolor{blue}{\hyperlink{ref:36}{36}}-\textcolor{blue}{\hyperlink{ref:38}{38}}] & Infeasible offspring & (1) NDS; (2) CD & Infeasible
parents & (1) NDS; (2) CD \\
RI2 & Infeasible offspring & (1) NDS; (2) CD & Infeasible parents & (1)
NDS; (2) ED \\
RI3 & Infeasible offspring and infeasible parents & (1) NDS; (2) CD &
Infeasible parents & (1) NDS; (2) ED \\
RI4 & Infeasible offspring and infeasible parents & (1) NDS; (2) CD &
Infeasible parents & ED \\
\end{longtable}
}

\begin{center}
\protect\phantomsection\label{tab:2}{}
Table 2. Repair strategies for partially infeasible populations (RF).
\end{center}

{\def\LTcaptype{none} 
\begin{longtable}[]{@{}
  >{\centering\arraybackslash}p{(\linewidth - 8\tabcolsep) * \real{0.1521}}
  >{\centering\arraybackslash}p{(\linewidth - 8\tabcolsep) * \real{0.2006}}
  >{\centering\arraybackslash}p{(\linewidth - 8\tabcolsep) * \real{0.2252}}
  >{\centering\arraybackslash}p{(\linewidth - 8\tabcolsep) * \real{0.2006}}
  >{\centering\arraybackslash}p{(\linewidth - 8\tabcolsep) * \real{0.2205}}@{}}
\toprule\noalign{}
\begin{minipage}[b]{\linewidth}\centering
Strategy
\end{minipage} & \begin{minipage}[b]{\linewidth}\centering
Candidates
\end{minipage} & \begin{minipage}[b]{\linewidth}\centering
Selection of candidates
\end{minipage} & \begin{minipage}[b]{\linewidth}\centering
Donors
\end{minipage} & \begin{minipage}[b]{\linewidth}\centering
Selection of donors
\end{minipage} \\
\midrule\noalign{}
\endhead
\bottomrule\noalign{}
\endlastfoot
RF1 [\textcolor{blue}{\hyperlink{ref:36}{36}}-\textcolor{blue}{\hyperlink{ref:38}{38}}] & Infeasible offspring & (1) NDS; (2) CD & Feasible
parents & (1) First front; (2) ED \\
RF2 & Infeasible offspring & (1) NDS; (2) CD & Feasible parents & ED \\
RF3 & Infeasible offspring & (1) NDS; (2) CD & Infeasible and feasible
parents & (1) First front; (2) ED \\
\end{longtable}
}

\subsection{Comparative runs with other
algorithms}\label{comparative-runs-with-other-algorithms}

To compare the performance and efficiency of the proposed optimization
algorithm, we have used the latest version of PlatEMO platform [\textcolor{blue}{\hyperlink{ref:39}{39}}].
PlatEMO is an open-source MATLAB-based platform designed specifically
for conducting reproducible and systematic experiments in evolutionary
multi-objective optimization. We first executed all constrained
multi-objective optimization algorithms available in the PlatEMO package
(43 algorithms) on the tanker design problem (details in Section 3.2)
for 10,000 function evaluations. The tanker problem is the most
difficult and time-consuming problem in this study. This initial
screening allowed us to identify which algorithms are capable of
producing feasible and competitive solutions within the evaluation
budget. Based on their performance, the 12 best-performing algorithms
were selected for more thorough analysis, applied to all test cases.

Beyond selecting algorithms based solely on empirical performance, this
comparison establishes a baseline against a diverse set of evolutionary
and nature-inspired multi-objective optimizers with varying
constraint-handling mechanisms, selection schemes, and search
strategies. This is particularly important for constrained problems with
many decision variables and nonlinear constraints, such as the tanker
problem considered in this study, where different algorithmic features
may be favored. By evaluating all algorithms across multiple test cases,
the comparison provides more statistically reliable insights and reduces
problem-specific bias, while also highlighting the robustness,
strengths, and limitations of the proposed repair mechanism relative to
well-established methods.

\section{Test problems}\label{test-problems}

The performance of our proposed algorithm is compared against other
algorithms on three complex engineering problems. These problems are
explained below.

\subsection{72-bar and 120-bar truss
problems}\label{bar-and-120-bar-truss-problems}

The 72-bar and 120-bar truss problems are well-known problems in
structural engineering [\textcolor{blue}{\hyperlink{ref:46}{46}}-\textcolor{blue}{\hyperlink{ref:49}{49}}]. The objectives are to minimize the
total structural mass and maximize stiffness, the latter represented by
compliance. The constraints are the difference between the allowable
stress limit in each member and the applied stresses. To further
increase the complexity of the test cases, we additionally incorporated
buckling constraints into both truss models. As a result, we have 144
constraints for the 72-bar truss and 240 constraints for the 120-bar
truss problem. For the 72-bar truss problem, two independent load cases
are considered. The maximum stress resulting from these two cases is
used to evaluate the structural safety of each member, thereby
increasing the difficulty of the optimization task. In Load Case 1, a
vertical force \emph{P} = (2·10\textsuperscript{6},
2·10\textsuperscript{6}, -2·10\textsuperscript{6}) N is applied at node
1, see \textcolor{blue}{\hyperref[_Ref217307988]{Fig. 2}}. In Load Case 2, a vertical
force \emph{P} = (0, 0, -2·10\textsuperscript{6}) N is applied to nodes
1, 2, 3, and 4. For the 120-bar truss problem, a mass-based benchmark
loading scaled by a factor of 20 is adopted. At node 38, the applied
force is \(F_{1}\) = (0, 0, -3000·20·\emph{g}) N\emph{,} for nodes 14
and 39-49, the applied force is \(F_{2}\) = (0, 0, -500·20·\emph{g})
N\emph{,} and all remaining nodes receive \(F_{3}\) = (0, 0,
-100·20·\emph{g}) N with \(g = 9.81\frac{m}{s^{2}}\).
\textcolor{blue}{\hyperref[_Ref217307988]{Fig. 2}} presents
schematic diagrams of both truss structures, including node numbering
and element connectivity.

\makebox[\linewidth][c]{\includegraphics[width=5.91667in,height=4.01461in]{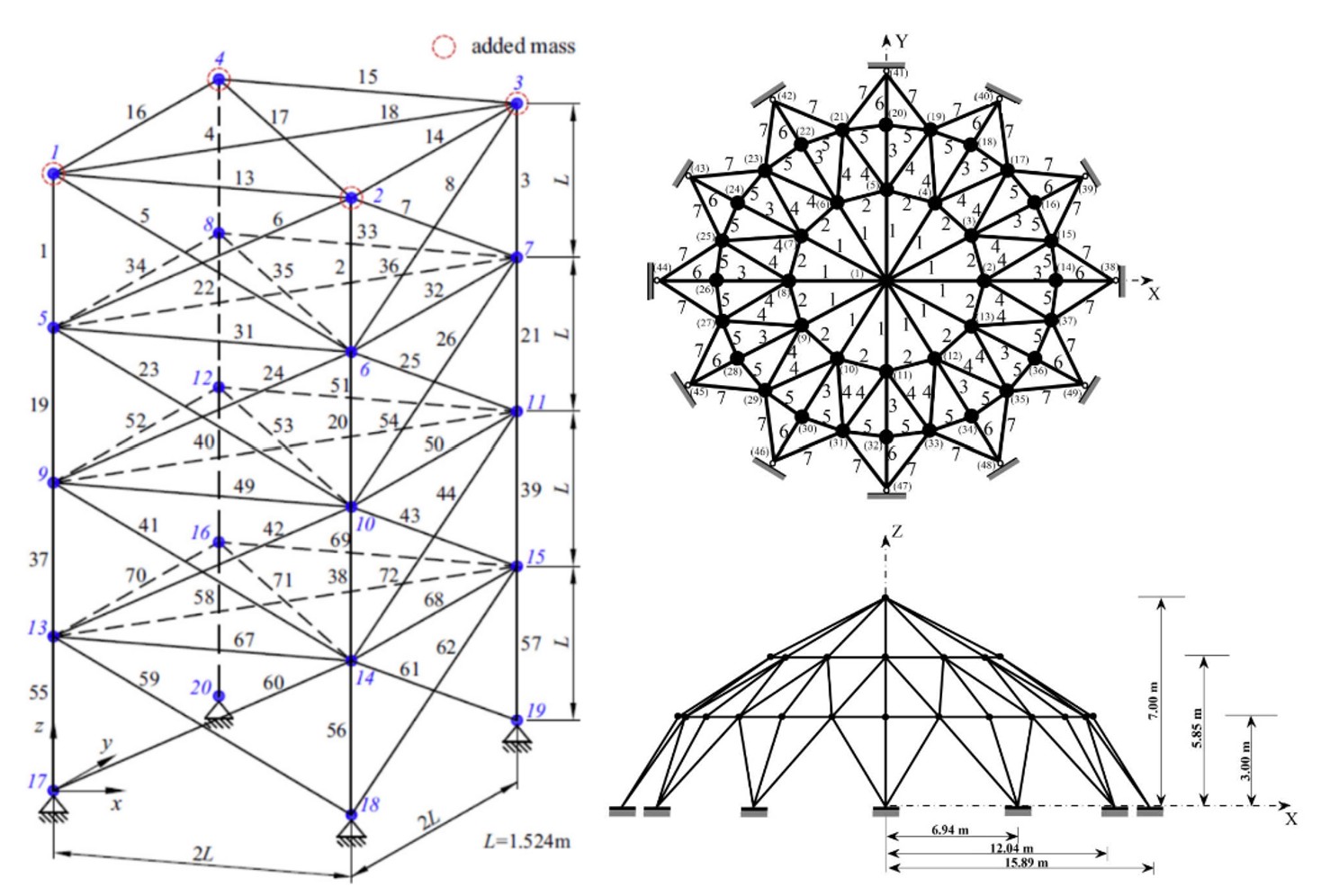}}

\begin{center}
\protect\phantomsection\label{_Ref217307988}{}Fig. 2. Schematic of
72-bar truss (left) and 120-bar truss (right) truss problems. For the
72-bar truss, blue numbers denote the nodes and black numbers denote the
members. For the 120-bar truss, numbers without parentheses indicate
members, while numbers in parentheses indicate nodes.
\end{center}

Young's modulus of the material is 200 GPa and the density is 7850
kg/m³. The design variables correspond to the cross-sectional areas of
the circular truss members, bounded between 0.001 m² and 0.021 m² for
both test problems. Buckling capacity is evaluated using Euler's
critical stress formula, \(\sigma_{cr} = \frac{\pi^{2}EI}{AL^{2}}\) and
the allowable normal stress for all members is set to 400 MPa.

\makebox[\linewidth][c]{\includegraphics[width=5.5316in,height=2.75143in]{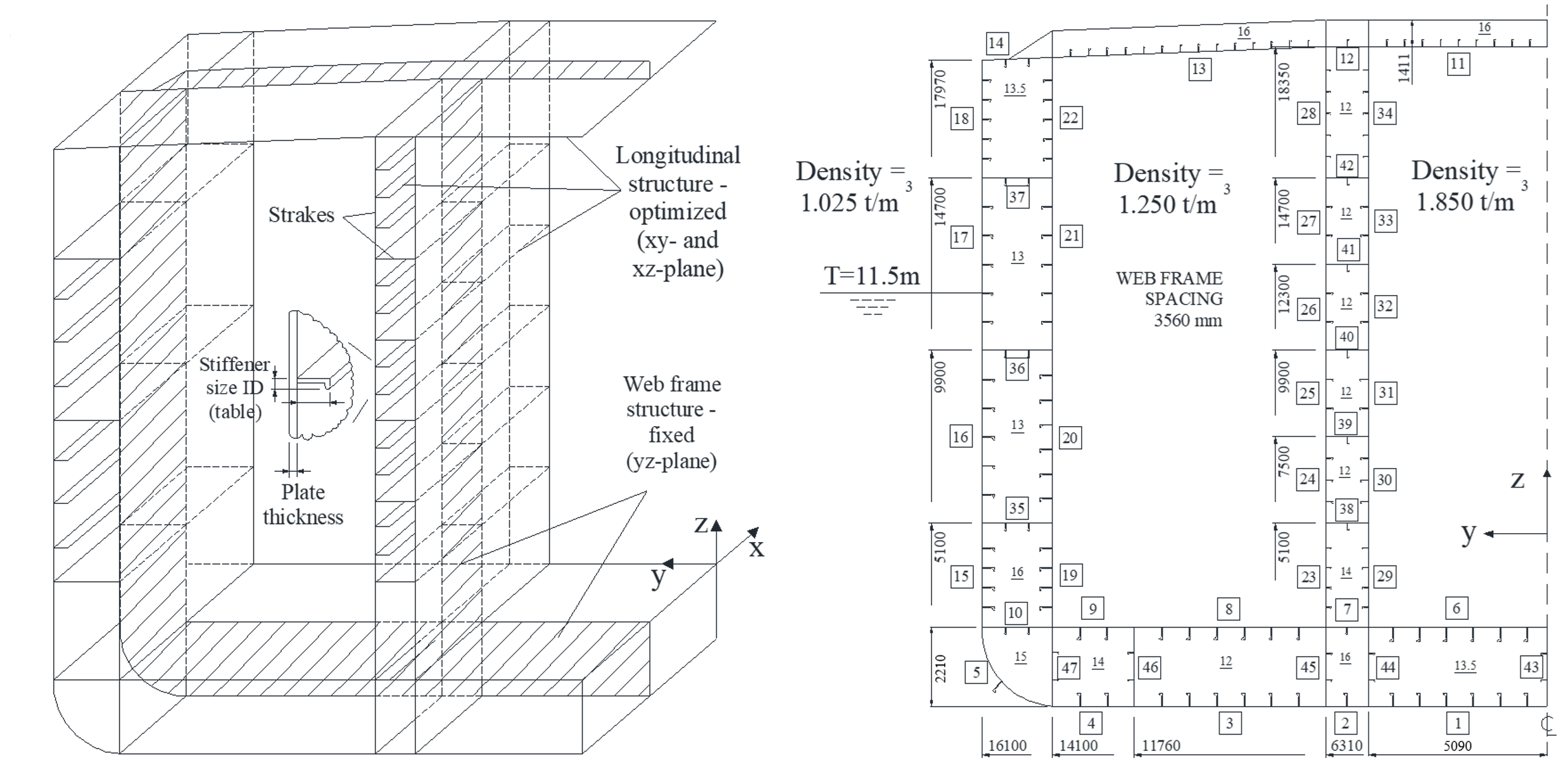}}

\begin{center}
\protect\phantomsection\label{_Ref215570211}{}Fig. 3. Structure of the tanker. The numbers in squares are panel identification (ID) numbers.
\end{center}

\subsection{Real-life optimization problem: Main frame of a ship
tanker}\label{real-life-optimization-problem-main-frame-of-a-ship-tanker}

As a practical engineering case study, we consider the structural
optimization of a chemical tanker. The vessel measures 180 m in length,
32 m in breadth, and 18 m in depth, with a draught of 11.5 m, and
operates under standard service conditions. A schematic of the
structural layout is provided in \textcolor{blue}{\hyperref[_Ref215570211]{Fig. 3}}. The
optimization focuses on the longitudinal structural members, and the
model is implemented using MATLAB and Fortran. This problem was
originally introduced in [\textcolor{blue}{\hyperlink{ref:50}{50}}] and later revisited and optimized in
[\textcolor{blue}{\hyperlink{ref:36}{36}}, \textcolor{blue}{\hyperlink{ref:51}{51}}]. The formulation includes 94 variables and 376 nonlinear
structural failure constraints. The associated variables, structural
constraints, and objective functions are detailed below. Previous
studies have shown that genetic algorithm-based optimization of
large-scale marine vessels requires a substantial number of function
evaluations, resulting in high computational costs [\textcolor{blue}{\hyperlink{ref:52}{52}}, \textcolor{blue}{\hyperlink{ref:53}{53}}]. Several
methods have been proposed to mitigate this issue, although with limited
success [\textcolor{blue}{\hyperlink{ref:54}{54}}].

\subsubsection{Design variables and
parameters}\label{design-variables-and-parameters}

The tanker is subjected to lateral pressures from both the sea and the
cargo. Variations in the local balance between buoyancy and the ship's
weight generate vertical shear forces and bending moments along the hull
under both sagging and hogging conditions [\textcolor{blue}{\hyperlink{ref:55}{55}}]. The maximum bending
moments reach 2.93 × 10⁶ kN.m in sagging and 2.41 × 10⁶ kN.m in hogging,
while the maximum vertical shear force is 48 × 10³ kN in both
conditions.

Due to structural symmetry, only half of the midship section is modeled
in the optimization (see \textcolor{blue}{\hyperref[_Ref215570211]{Fig. 3}}), although all
reported masses in the results correspond to the full structure. The
model contains 47 panels, each characterized by two design parameters:
the plate thickness and the stiffener identification number, the later
selected from a continuous table featuring stiffener's height and
thickness. This results in 94 design variables in total. The stiffener
type, the number of stiffeners, the material properties, and all
transverse structural components are kept constant throughout the
optimization. Plate thickness values range from 5 mm to 26 mm in 1-mm
increments. Stiffener profiles are selected from standard steel-producer
catalogs (e.g., [\textcolor{blue}{\hyperlink{ref:56}{56}}]), covering the range from HP 100×5 to HP
430×17.

\subsubsection{Design constraints}\label{design-constraints}

The cargo tank plating has a yield strength of 460 MPa, while the
remainder of the structure uses a material with a yield strength of 355
MPa. Each panel is subjected to eight structural constraints: stiffener
yielding, plate yielding, plate buckling, stiffener web buckling,
stiffener flange buckling, lateral buckling, tripping, and the crossover
constraint, resulting in 376 constraints in total. The crossover
constraint prevents uncontrolled panel collapse by ensuring that the
global buckling strength exceeds both plate and stiffener buckling
strengths [\textcolor{blue}{\hyperlink{ref:57}{57}}]. Stresses generated by global loads (shear force and
bending moment) are computed using the Coupled Beam (CB) method [\textcolor{blue}{\hyperlink{ref:58}{58}}]
for both sagging and hogging conditions across the cross-section. Local
stresses in plates and stiffeners due to lateral pressure are evaluated
using the classic plate and beam theory. These local stresses are then
superimposed on the global stresses, and the most critical combination
is used for each panel. At last, all constraints are scaled using the
nonlinear normalization scheme from [\textcolor{blue}{\hyperlink{ref:59}{59}}]:

\begin{equation}
g_{j}\left(\mathbf{x}\right)
= \frac{A_{j}\left(\mathbf{x}\right)
        - \left|B_{j}\left(\mathbf{x}\right)\right|}
       {A_{j}\left(\mathbf{x}\right)
        + \left|B_{j}\left(\mathbf{x}\right)\right|}.
\tag{2}
\end{equation}

where \(A_{j}\left( \mathbf{x} \right)\) is the capacity of structural
element \emph{j}, and \(B_{j}\left( \mathbf{x} \right)\) is the
corresponding acting stress. The normalized constraint values lie
between --1 and 1, with negative values indicating constraint violation
and zero marking the allowable limit.

\subsubsection{Objective functions}\label{objective-functions}

Two objective functions are considered: (i) minimization of the
structural mass and (ii) maximization of the deck adequacy. The mass is
computed by multiplying the steel density (8 t/m³) by the
cross-sectional areas of the longitudinal members and extending them
along the full length (and breadth) of the vessel, effectively treating
the hull as a prismatic structure. In addition, 21.44 t is included as
the mass of the transverse structure for every 3.56 m of ship length.
Deck adequacy is used as a simplified safety indicator and is defined as
the sum of the normalized deck-related constraints. An increase in deck
adequacy effectively reduces stresses in the deck, which is a critical
structural component due to its single-plated construction, in contrast
to the double-bottom and double-side regions. The two objectives are
inherently conflicting: lighter designs tend to experience higher
stresses and therefore exhibit lower adequacy. Conversely, increasing
plate and stiffener thicknesses reduces deck stresses but results in a
heavier structure. The second objective therefore promotes safer designs
by encouraging increased deck stiffness at the expense of additional
mass. The exact relationship between deck stress and structural
dimensions is complex, as the surrounding structure also influences the
second moment of area of the cross-section. Moreover, local loads vary
between compartments, and the resulting stresses are superimposed on the
stresses induced by global loading.

\subsubsection{Surrogate model for tanker problem
evaluation}\label{surrogate-model-for-tanker-problem-evaluation}

Evaluating each design using the CB method is computationally expensive,
requiring approximately 4 seconds on a standard desktop computer. This
cost becomes prohibitive given that thousands of designs must be
assessed in each optimization run. To mitigate this expense, a
feed-forward neural network based on a multilayer perceptron is employed
as a surrogate model to predict stresses and, consequently, the
corresponding constraint values. The first objective, structural mass,
is computed directly from the decision variables and the steel density.
The second objective is derived from the constraint values, as it
represents the sum of the normalized deck-related constraints (as
explained in Section \hyperref[objective-functions]{3.2.3}).

The MLP was trained using 20,000 datasets, with 80\% of the samples
allocated for training and the remaining 20\% used for testing. To
ensure accurate prediction of constraint values, a separate surrogate
model was trained for each constraint, while maintaining the same
network architecture across all models. The network architecture is
illustrated in \textcolor{blue}{\hyperref[_Ref215577822]{Fig. 4}}. Training was conducted
using the Adam optimizer, which is particularly effective for problems
with sparse gradients and adaptive learning rates. The training process
spanned 50 epochs, with a mini-batch size of 64 samples to balance
memory efficiency and gradient estimation stability. To prevent the
model from learning spurious correlations related to the data order, the
training data were shuffled at the start of each epoch. A piecewise
learning rate schedule was employed, reducing the learning rate by a
factor of 0.5 every 20 epochs, facilitating smoother convergence and
reducing the risk of overshooting minima. The trained networks allow
reduction of optimization time for each run to about four hours on a
standard desktop computer.

The trained surrogate model achieves an accuracy of 99.05\%, computed as 
\(1 - \widetilde{e}\), where \(\widetilde{e}\) is the median relative
error evaluated on the test dataset. This high level of accuracy
demonstrates that the surrogate is sufficiently reliable to replace the
original structural model for design evaluation, leading to a
substantial reduction in computational cost.

\makebox[\linewidth][c]{\includegraphics[width=5.02941in,height=1.60905in]{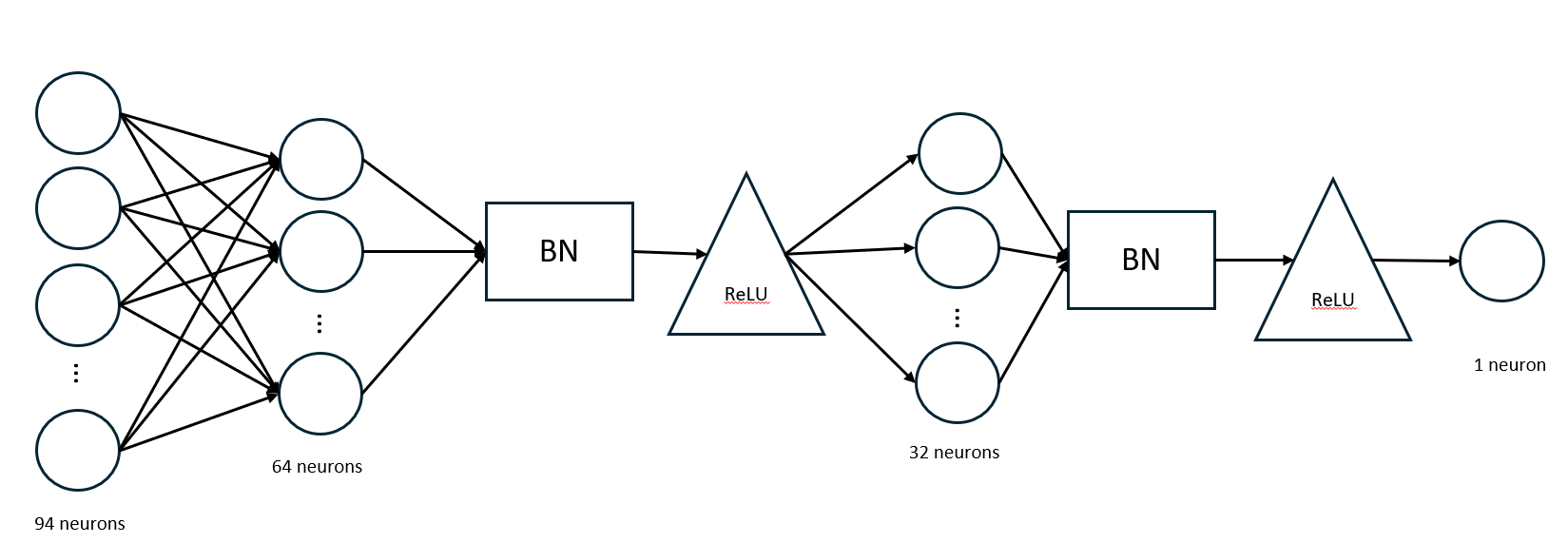}}

\begin{center}
\protect\phantomsection\label{_Ref215577822}{}Fig. 4. Architecture of
the surrogate model to evaluate each constraint of the tanker problem.
\end{center}

\subsection{Performance measures}\label{performance-measures}

The primary objective in multi-objective optimization is to obtain a
nondominated solution set that both converges closely to the true PF and
exhibits good diversity. To assess convergence performance in this
study, the inverted generational distance (IGD) metric is employed. IGD
[\textcolor{blue}{\hyperlink{ref:60}{60}}] quantifies the average Euclidean distance from each point on
the true PF to its nearest solution in the algorithm's nondominated set,
see \textcolor{blue}{\hyperref[fig:5]{Fig. 5}}. Only feasible solutions are considered in the IGD
calculation, with lower IGD values indicating better convergence.

For the truss benchmark problems, the true PF was constructed by running
the 12 best-performing algorithms from the PlatEMO package 20 times for
500 generations on each problem, followed by unifying the results and
extracting an evenly distributed set of nondominated solutions. For the
tanker problem, the same procedure was applied, but with 10 runs per
algorithm due to the substantially higher computational cost.

\makebox[\linewidth][c]{\includegraphics[width=1.9in,height=1.9in]{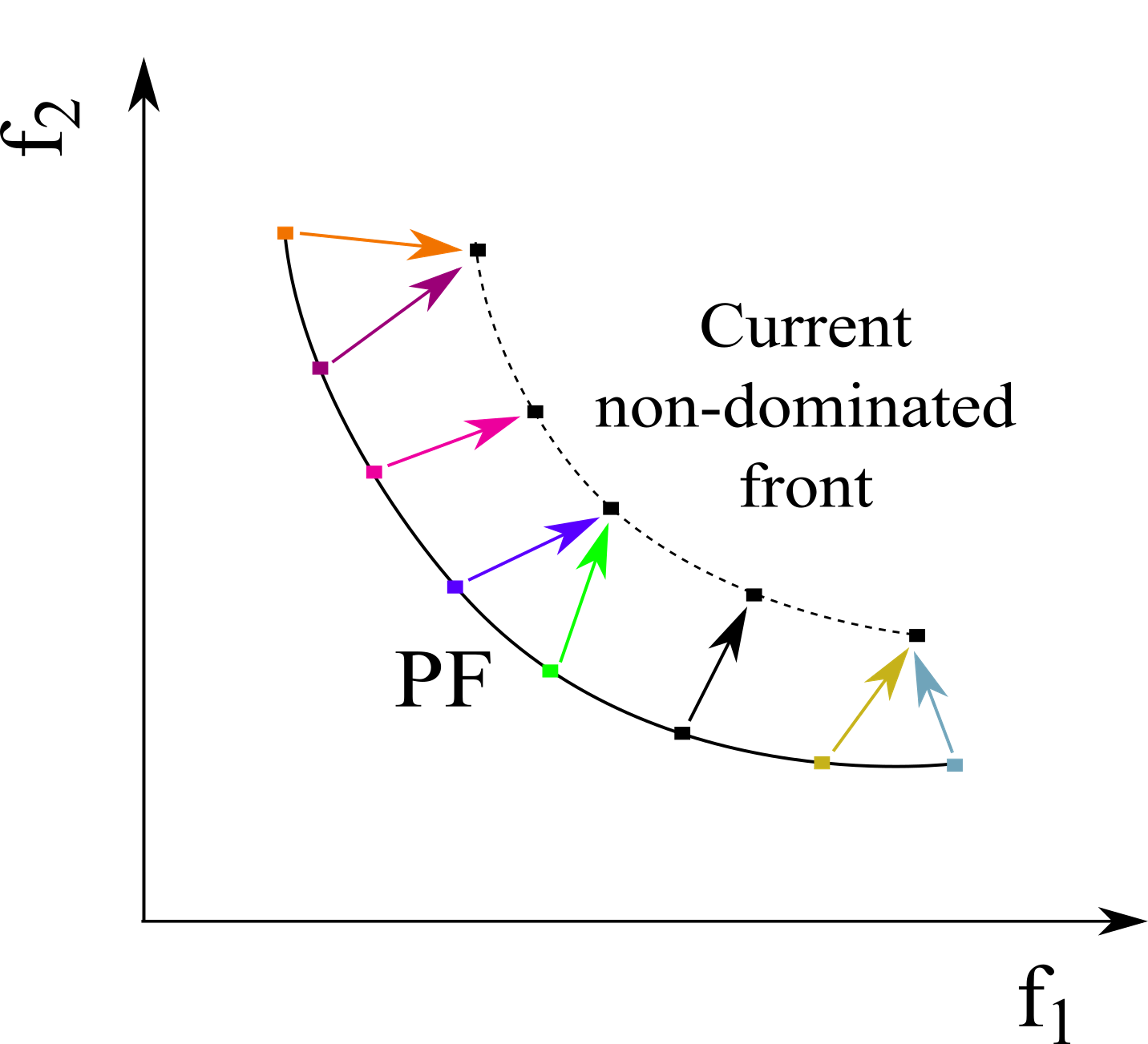}}
\begin{center}
\protect\phantomsection\label{fig:5}{}
Fig. 5. IGD calculation principle.
\end{center}

\section{Results and discussion}\label{results-and-discussion}

\subsection{Optimization comparisons}\label{optimization-comparisons}

First, all 43 algorithms suitable for constrained multi-objective
optimization problems (CMOPs) in the latest PlatEMO package [\textcolor{blue}{\hyperlink{ref:39}{39}}]
were run five times for 100 generations on the tanker problem, which
represents the most challenging test case considered in this study.
Based on these preliminary evaluations, the 12 best-performing
algorithms were selected according to their average final IGD values.
Subsequently, these selected algorithms, together with DPCME, were
executed on the full set of benchmark problems for 500 generations, with
a population size fixed at 100 for all runs. Each algorithm was run 20
times on the truss problems and 10 times on the tanker problem, to
ensure statistical reliability. As shown in
\textcolor{blue}{\hyperref[_Ref217306905]{Fig. 6}} for the 120-bar truss problem, the IGD
convergence curves obtained over 1000 generations indicate that the
first 500 generations are representative of the overall optimization
behavior, as convergence trends remain essentially unchanged thereafter.
Therefore, considering computational cost, IGD performance is reported
over 500 generations in the subsequent figures.

\makebox[\linewidth][c]{\includegraphics[width=3.10in,height=2.5in]{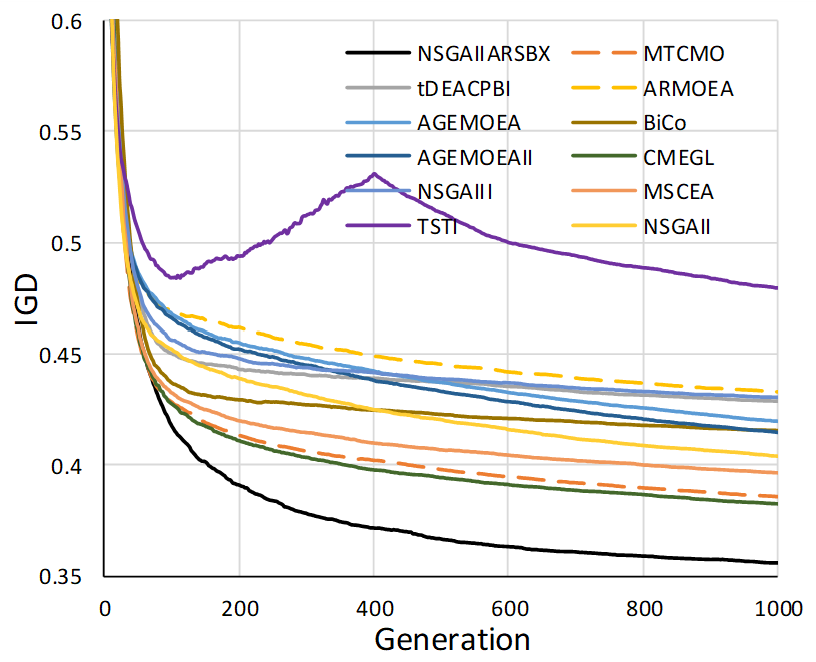}}

\begin{center}
\protect\phantomsection\label{_Ref217306905}{}Fig. 6. Mean
per-generation IGD (averaged over 20 runs) across 1000 generations for
the 12 best-performing PlatEMO algorithms on the 120-bar truss problem.
\end{center}

\textcolor{blue}{\hyperref[_Ref215182639]{Fig. 7}} presents the average IGD values of the
12 best-performing algorithms from the PlatEMO package across all test
problems. As can be observed, MTCMO achieves the best result on the
72-bar truss problem, second best on the tanker problem and third best
on the 120-bar truss problem. Meanwhile, NSGAIIARSBX achieves the best
performance on the 120-bar truss and tanker problems, outperforming
MTCMO by approximately 30\% on the tanker problem. NSGAIIARSBX is
surpassed by three other algorithms on the 72-bar truss problem.

\protect\phantomsection\label{_Ref215182639}{}\makebox[\linewidth][c]{\includegraphics[width=6.5in,height=5.80903in]{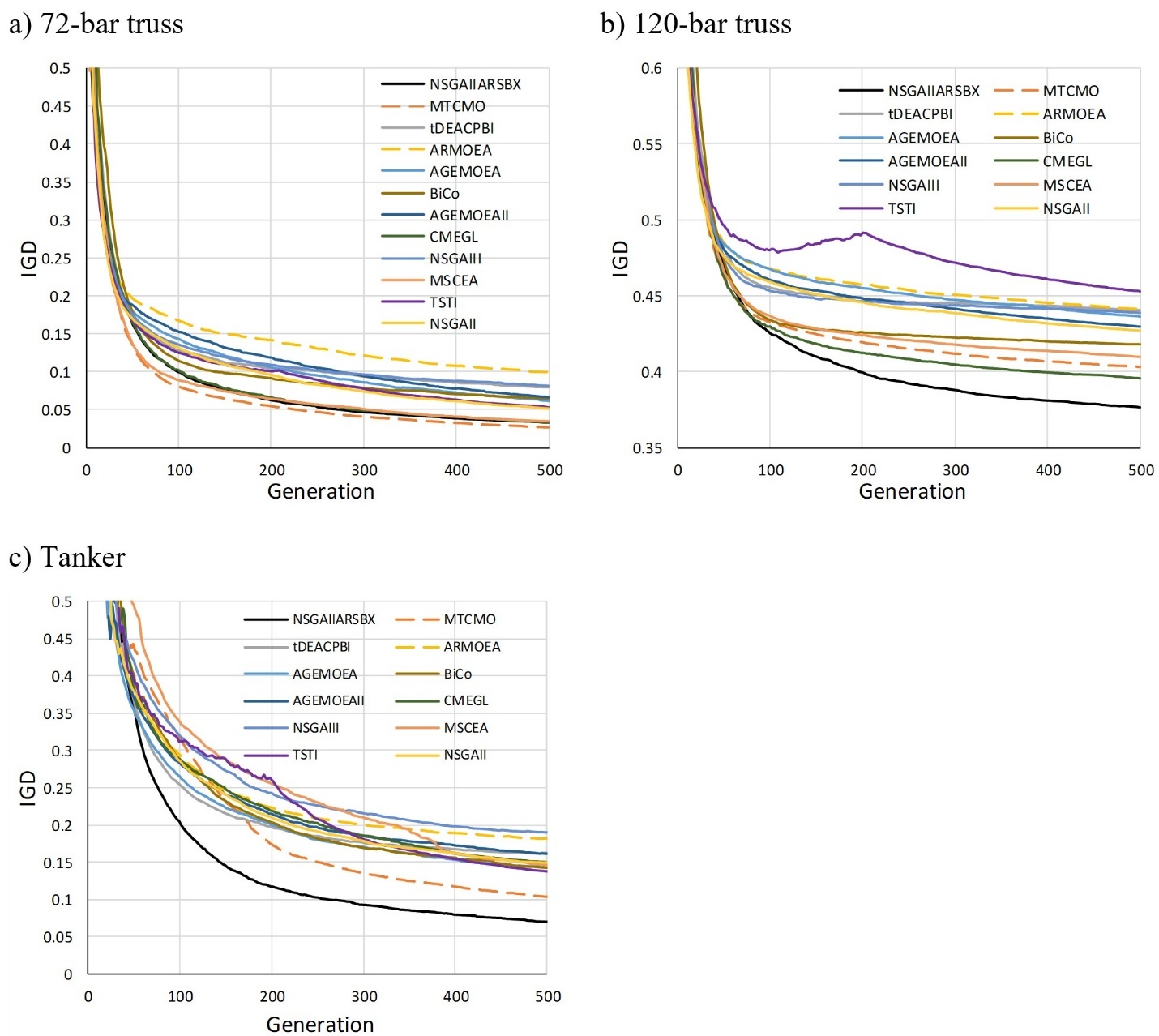}}

\begin{center}
Fig. 7. Average IGD of the 12 best-performing algorithms from the
PlatEMO platform on the three test cases examined in this study.
\end{center}

To evaluate different repair strategies for fully infeasible populations
(RI), we calculated the average IGD metric after the first generation of
the optimization process, since all test cases begin producing feasible
solutions immediately as the RI are applied. As shown in
\textcolor{blue}{\hyperref[_Ref215583982]{Table 3}}, the RI4 strategy achieves the best
IGD values for the 72-bar truss and tanker problems, and ranks second,
with only a slight difference, for the 120-bar truss problem.
\protect\phantomsection\label{_Ref215583982}{}Although the improvement
over RI1, which was proposed previously, is not large, RI4 consistently
shows better performance, which can be beneficial for enabling faster
progress of the optimization in subsequent generations. RI3 and RI4
exploit not only the initial (random) solutions but also solutions
generated in the subsequent generation using the baseline DPCME (set to
50\% of the total population), as seen in \textcolor{blue}{\hyperref[tab:1]{Table 1}}. Moreover, the use of
ED (either exclusively in RI4 or in combination with NDS in RI3),
instead of NDS and CD as used in RI1, may promote more successful
repairs at this early optimization stage, as neighboring designs tend to
have more similar genomes. Given that RI4 exhibits the best overall
performance, it is adopted for all subsequent runs employing RF-based
approaches across all test problems.

\begin{center}
Table 3. IGD values after applying the repair on fully infeasible
(initial) population (RI).
\end{center}

{\def\LTcaptype{none} 
\begin{longtable}[]{@{}
  >{\centering\arraybackslash}p{(\linewidth - 8\tabcolsep) * \real{0.2006}}
  >{\centering\arraybackslash}p{(\linewidth - 8\tabcolsep) * \real{0.1347}}
  >{\centering\arraybackslash}p{(\linewidth - 8\tabcolsep) * \real{0.1347}}
  >{\centering\arraybackslash}p{(\linewidth - 8\tabcolsep) * \real{0.1347}}
  >{\centering\arraybackslash}p{(\linewidth - 8\tabcolsep) * \real{0.1347}}@{}}
\toprule\noalign{}
\begin{minipage}[b]{\linewidth}\centering
\end{minipage} & \begin{minipage}[b]{\linewidth}\centering
RI1
\end{minipage} & \begin{minipage}[b]{\linewidth}\centering
RI2
\end{minipage} & \begin{minipage}[b]{\linewidth}\centering
RI3
\end{minipage} & \begin{minipage}[b]{\linewidth}\centering
RI4
\end{minipage} \\
\midrule\noalign{}
\endhead
\bottomrule\noalign{}
\endlastfoot
72-bar truss & 0.420 & 0.423 & 0.419 & 0.415 \\
120-bar truss & 0.648 & 0.648 & 0.641 & 0.642 \\
Tanker & 0.520 & 0.509 & 0.511 & 0.504 \\
\end{longtable}
}

\textcolor{blue}{\hyperref[_Ref215182641]{Fig. 8}} compares the best-performing algorithm
identified in \textcolor{blue}{\hyperref[_Ref215182639]{Fig. 7}} with DPCME and DPCME
combined with the best RF repair strategy, and also includes NSGA-II as
a baseline. NSGA-II is considered because it is still the most widely
used global optimization algorithm in engineering [\textcolor{blue}{\hyperlink{ref:61}{61}}] and has shown
remarkable performance on structural and ship optimization problems
[\textcolor{blue}{\hyperlink{ref:37}{37}}, \textcolor{blue}{\hyperlink{ref:38}{38}}]. For the 72-bar truss problem, DPCME outperforms NSGA-II
but is surpassed by MTCMO. However, DPCME-RF3, i.e., the proposed
algorithm with the best proposed repair strategy for this test problem
achieves about the same performance as MTCMO in the end of the
optimization, and is even better than MTCMO at early stages. For the
120-bar truss problem, DPCME again outperforms NSGA-II by a wide margin,
and the RF3 repair strategy further enhances its performance;
nevertheless, NSGAIIARSBX still achieves the best results in the end. A
similar behavior is observed for the tanker problem. Although
NSGAIIARSBX achieves an advantage for the latter two problems, the
improvement gained by integrating the repair strategy into DPCME is
substantial, reducing the performance gap with NSGAIIARSBX and
outperforming all at the beginning of the optimization.

\protect\phantomsection\label{_Ref215182641}{}\makebox[\linewidth][c]{\includegraphics[width=6.5in,height=5.85in]{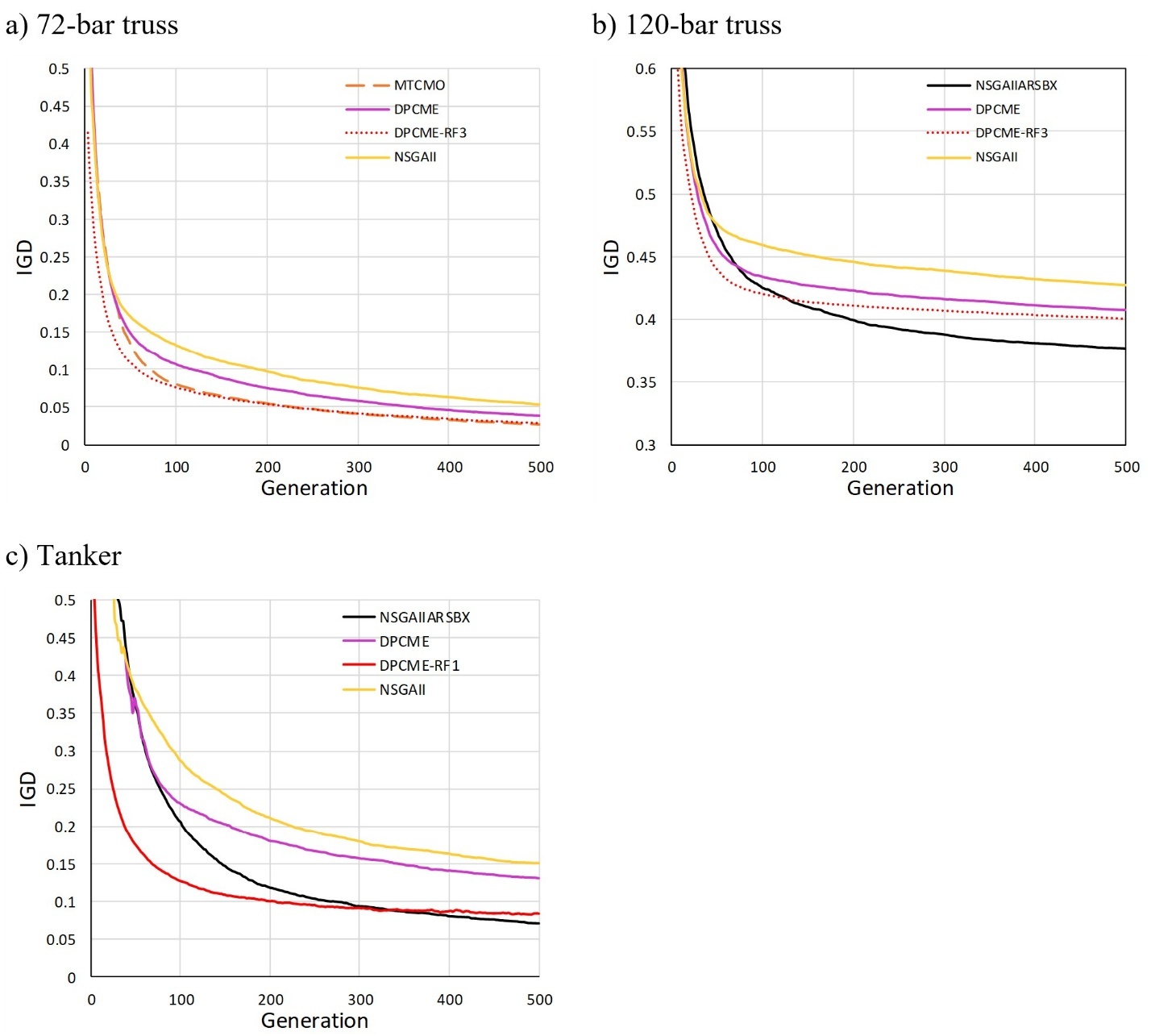}}

\begin{center}
Fig. 8. Average IGD for the best algorithm in the PlatEMO platform for
each case study, the benchmark NSGA-II, the proposed DPCME, and the
DPCME coupled with the best repair method.
\end{center}

\textcolor{blue}{\hyperref[_Ref215182969]{Fig. 9}} illustrates the performance of DPCME
without repair and with the three RF repair strategies, alongside the
best-performing algorithm from the PlatEMO package for reference.
Although the differences among the repair strategies are relatively
small, each of them significantly improves the performance of DPCME
compared to the adaptive threshold implemented as the only CHT in the
baseline DPCME. Moreover, even slight performance gains between repair
strategies can translate into substantial benefits in real-world
engineering applications, where small reductions in cost, weight, or
other design metrics are highly valuable. Overall, RF3 emerges as the
most effective repair strategy, with the only exception being the tanker
problem, where it is narrowly outperformed by RF1. RF3 differs from RF1
in that it also allows infeasible solutions to act as donors, as can be
seen in \textcolor{blue}{\hyperref[tab:2]{Table 2}}. This can lead to modest performance improvements in
some cases, as infeasible solutions may have better performance in
objective space. As demonstrated by the RI strategies, infeasible
solutions are capable of producing feasible solutions in a single repair
step. At later stages of the optimization, infeasible solutions
typically exhibit smaller overall constraint violations than at the
beginning of the search, which increases the likelihood of generating
feasible solutions.

\protect\phantomsection\label{_Ref215182969}{}\makebox[\linewidth][c]{\includegraphics[width=6.5in,height=5.81944in]{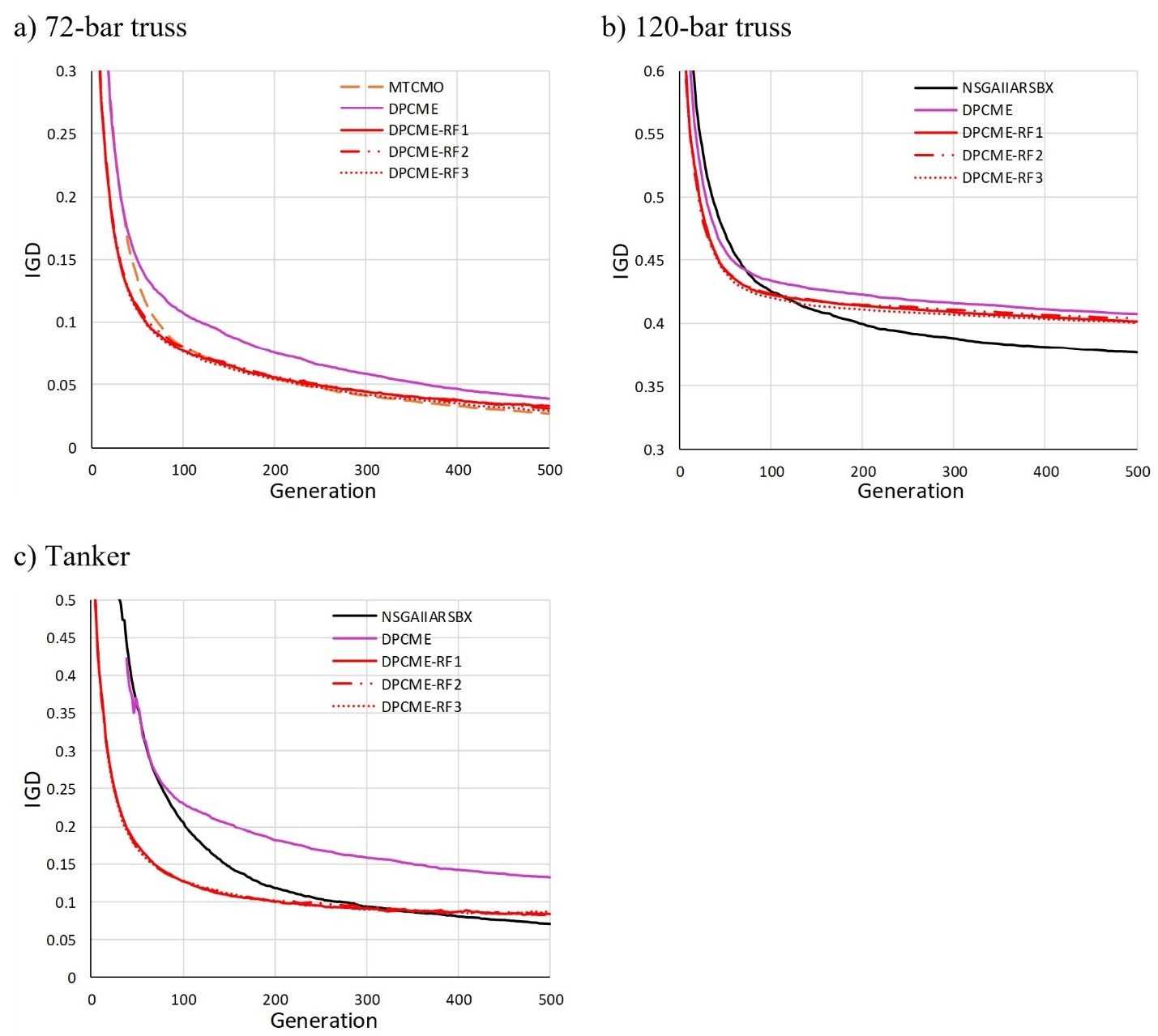}}

\begin{center}
Fig. 9. Average IGD of the base DPCME and its variants with second-stage
repair strategies assessed in this study, compared with the
best-performing algorithm from the PlatEMO platform.
\end{center}

\textcolor{blue}{\hyperref[_Ref215186919]{Fig. 10}} presents box plots of the final
population's IGD values for the best-performing PlatEMO algorithm,
NSGAII (as a benchmark), DPCME, and DPCME with the best repair strategy,
evaluated over 20 runs for the 72-bar truss problem and 10 runs for the
tanker problem. Box plots are a convenient way to illustrate the
distribution of final IGD values across multiple runs, allowing us to
assess both central tendency and variability of algorithm performance.
For the 72-bar truss case, both MTCMO and DPCME-RF3 exhibit compact box
and whisker ranges centered around low IGD values, indicating not only
strong performance but also high run-to-run reliability. In contrast,
NSGAII and baseline DPCME show noticeably wider boxes, reflecting
greater variability and less consistent behavior across independent
runs. NSGAII in particular produces substantially higher IGD values,
confirming its inferior performance relative to the proposed methods.
For the tanker problem, NSGAII displays noticeable variation across
runs, whereas the other algorithms maintain more consistent behavior. In
particular, NSGAIIARSBX and DPCME-RF1 produce both low IGD values and
narrow distribution ranges, highlighting their stable and reliable
performance. A similar behavior is observed for the 120-bar truss
problem; therefore, it is omitted here for brevity.

\protect\phantomsection\label{_Ref215186919}{}\makebox[\linewidth][c]{\includegraphics[width=6.5in,height=2.78194in]{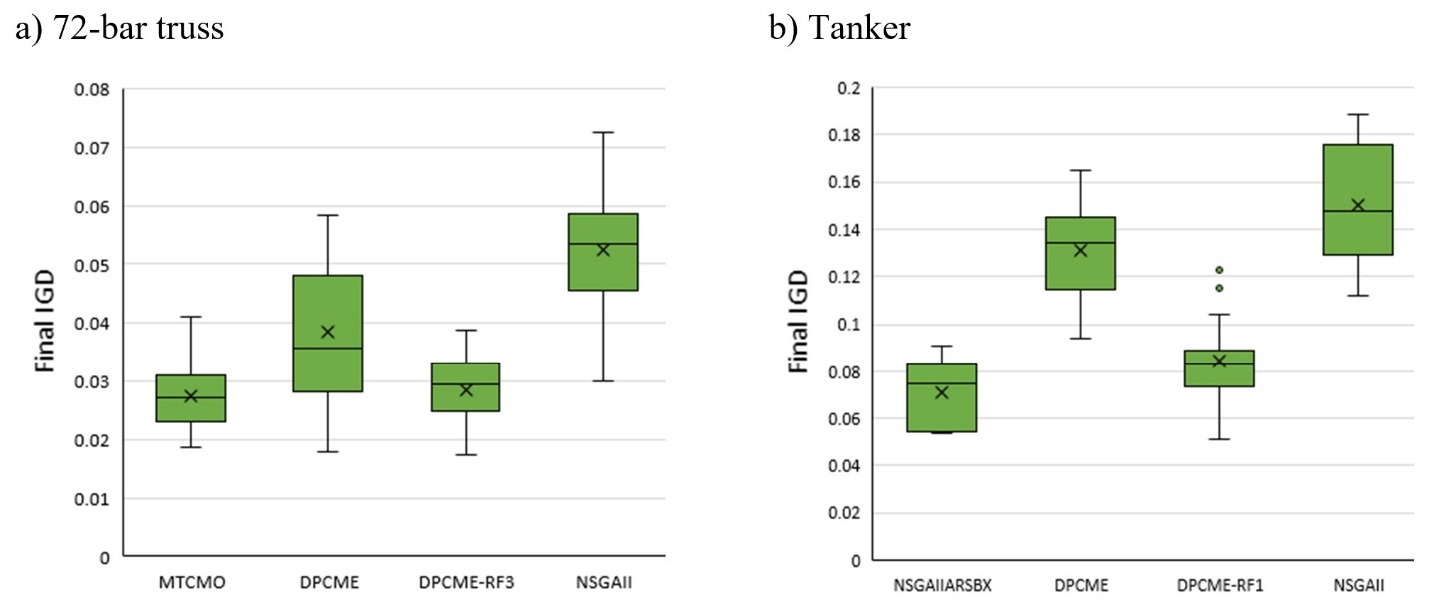}}

\begin{center}
Fig. 10. Box plots for the final population for all 10 runs for tanker
and 20 runs for 72-bar truss problem.
\end{center}

\textcolor{blue}{\hyperref[_Ref215190250]{Fig. 11}} shows the average number of solutions
belonging to different categories of solutions based on feasibility and
non-dominance, averaged over all runs for each test case. The plots on
the left-hand side present the total number of infeasible designs and
the subset that are non-dominated − these form the pool of repair
candidates. Across all problems, we observe a consistent trend: the
number of infeasible designs increases as the optimization progresses.
This occurs because, as the population converges toward the feasibility
boundaries, small perturbations from evolutionary operators (e.g.,
mutation and crossover) are increasingly likely to push solutions
outside the feasible region. In effect, the search becomes more
sensitive near constraint boundaries, leading to a higher proportion of
infeasible offspring in later generations. This rise in infeasible
solutions creates an opportunity for the repair mechanism to have
significant impact. The repair method can exploit these infeasible
designs to generate high-quality feasible solutions. For the 72-bar
truss and tanker problems, the number of non-dominated infeasible
designs also increases over time, while for the 120-bar truss problem
this number remains roughly constant. The plots on the right of the
figure summarize the performance of the repair procedure, showing the
number of repaired solutions, the portion that become feasible, and the
subset of those feasible designs that are non-dominated. For the 72-bar
truss and 120-bar truss problems, nearly all feasible repaired solutions
are non-dominated, demonstrating the strong effectiveness of the repair
method. Even for the more complex tanker problem, most feasible repaired
solutions still lie on the non-dominated front. The specified repair
rate is fixed at 10\% of the population, following detail study in
[\textcolor{blue}{\hyperlink{ref:36}{36}}], which explains why the number of repaired solutions eventually
stabilizes at that level for the tanker. Importantly, although the
fraction of the population that becomes successfully repaired (i.e.,
non-dominated feasible repaired solutions) is small, it nonetheless
leads to substantial performance improvements over the baseline DPCME
algorithm, as demonstrated above.

\protect\phantomsection\label{_Ref215190250}{}\makebox[\linewidth][c]{\includegraphics[width=6.1in,height=8.0in]{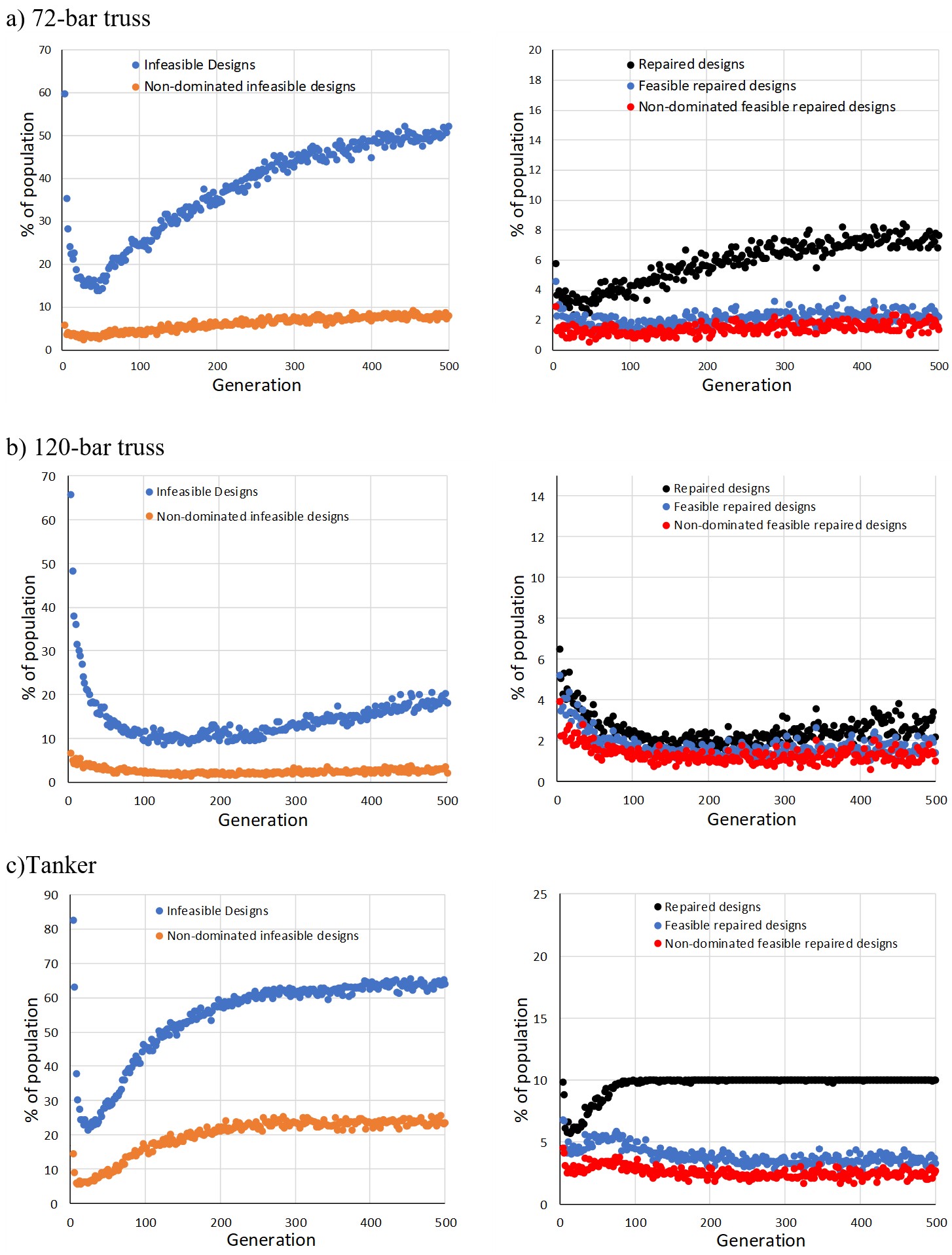}}

\begin{center}
Fig. 11. Average number of different categories of solutions over all
runs of the best repair strategy.
\end{center}

\protect\phantomsection\label{_Ref215194557}{}\makebox[\linewidth][c]{\includegraphics[width=6.5in,height=5.85694in]{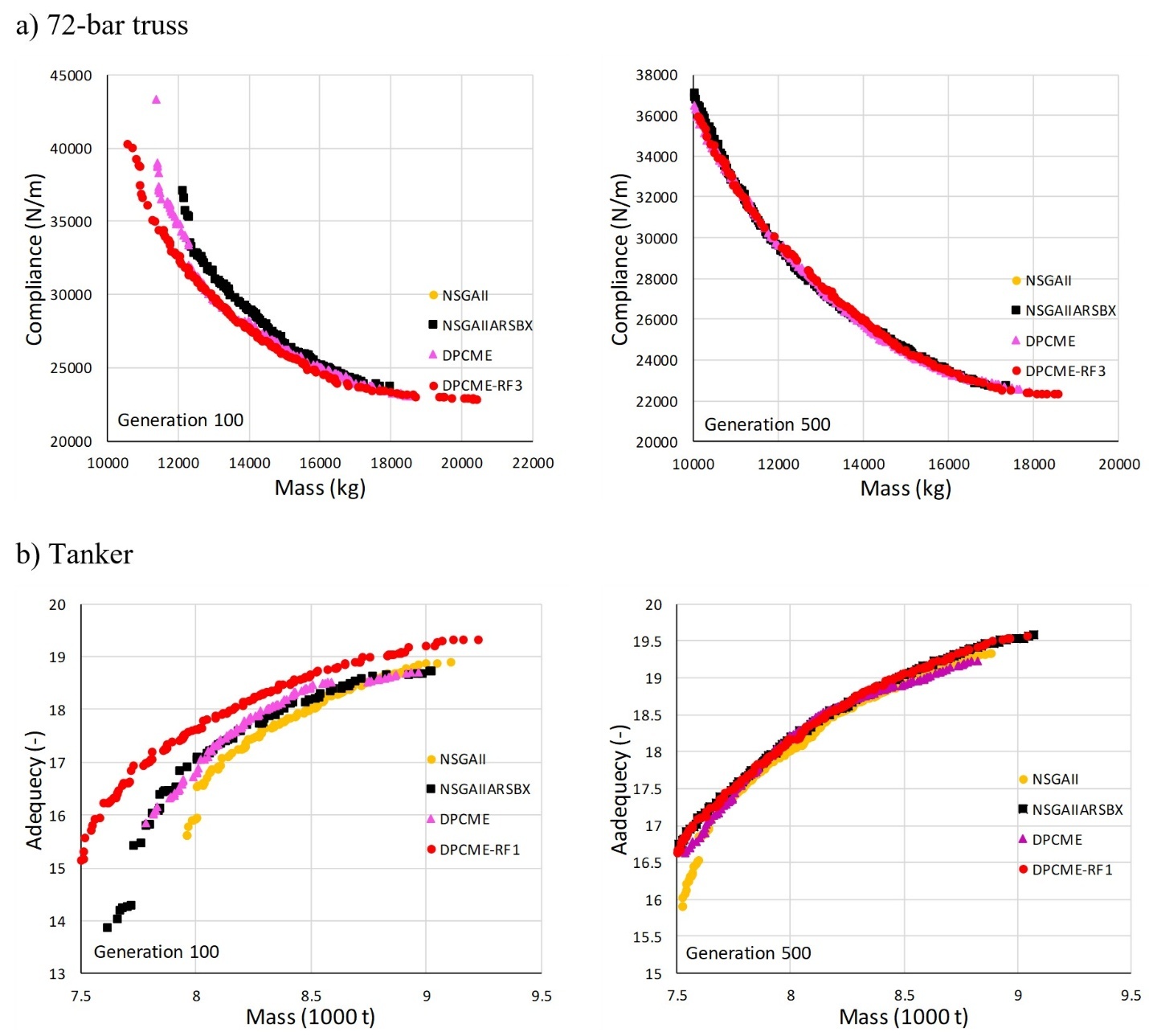}}

\begin{center}
Fig. 12. Final (right) and generation 100 (left) non-dominated fronts of
the best performing algorithm from PlatEMO package, NSGAII benchmark
algorithm, DPCME, and DPCME with the best repair strategy on a) 72-bar
truss and b) Tanker problems.
\end{center}

\textcolor{blue}{\hyperref[_Ref215194557]{Fig. 12}} presents the non-dominated fronts
obtained at generation 100 (left) and at the final generation (right).
These fronts were filtered from all the runs using non-domination
sorting. In many practical situations, an optimization algorithm may not
be allowed to run for a large number of generations due to time
limitations. Therefore, showing the fronts at generation 100 allows us
to evaluate the early-stage performance of the algorithms. After 100
generations, DPCME combined with the best repair strategy clearly
outperforms the other competitors with a substantial margin and better
spread on both test cases. At the final generation, for 72-bar truss,
DPCME-RF3 performs very close to NSGAIIARSBX. It exhibits a superior
spread and dominates NSGAIIARSBX at many locations along the front. For
the tanker problem, DPCME-RF1 also achieves performance close to
NSGAIIARSBX, although NSGAIIARSBX maintains a slightly better continuity
for higher mass values. It is also worth noticing that DPCME-RF1
noticeably outperforms NSGAII and the baseline DPCME. Similar trends are
observed for 120-bar truss problem, which is omitted here for brevity.

As observed from the results, both MTCMO and NSGAII-ARSBX demonstrate
strong performance on the considered test cases. To better understand
the factors contributing to their strong performance, we examine the
main work principles of these algorithms. MTCMO employs a
dual-population framework with explicit knowledge transfer between a
main and an auxiliary task, similar in principle to the proposed
approach. The main task relies on a Pareto-focused environmental
selection strategy, comparable to that used in DPCME, while the
auxiliary task adopts a dynamic ε-based constraint-handling technique,
which likely contributes to its strong performance on constrained
engineering problems [\textcolor{blue}{\hyperlink{ref:62}{62}}]. In contrast, NSGAII-ARSBX is a
single-population algorithm that follows the standard NSGA-II framework
but introduces a modified offspring generation mechanism. It combines
conventional simulated binary crossover (SBX) with a rotated SBX
operator, where crossover is performed in a PCA-rotated decision space
derived from the covariance structure of the current population. The
algorithm adaptively controls the proportion of offspring generated by
each crossover operator based on their relative success in producing
high-quality solutions during the optimization process [\textcolor{blue}{\hyperlink{ref:63}{63}}].

\subsection{Non-parametric tests}\label{non-parametric-tests}

To provide additional comparison of algorithms' performance, a
non-parametric statistical test was conducted. Among such tests, the
Friedman test is one of the most widely used. It is a rank-based test
that evaluates whether significant differences exist among multiple
methods across repeated measurements. In this test, each method is
ranked for every run, with the best-performing method assigned rank 1,
the second-best rank 2, and so on. The ranks are then averaged over all
runs, producing the mean rank of each algorithm. Finally, the algorithms
can be sorted based on their mean ranks to provide an overall ranking.
\textcolor{blue}{\hyperref[_Ref215610686]{Table 4}} presents the mean ranks and overall
rankings of the algorithms on all test cases, based on the final IGD
values. On the 72-bar truss problem, MTCMO achieves the best rank,
slightly ahead of DPCME-RF3. The other repair strategies follow, with
NSGAIIARSBX ranked thereafter. These results are consistent with the
trends observed in \textcolor{blue}{\hyperref[_Ref215182639]{Fig. 7}}. For the 120-bar
truss problem, NSGAIIARSBX dominates, followed by the CMEGL algorithm.
Next come the various DPCME variants with repair, MTCMO algorithm, and
finally the baseline DPCME. For the tanker problem, NSGAIIARSBX leads,
followed by DPCME with different repair strategies. A notable gap exists
between the mean rank of the last DPCME repair variant and MTCMO, with
the baseline DPCME ranked after MTCMO.

The Wilcoxon test is a non-parametric statistical method used to compare
the performance of two algorithms over multiple runs. Unlike parametric
tests, it does not assume a normal distribution of the data, making it
well-suited for performance metrics such as IGD, which often have skewed
distributions. In this study, we apply the Wilcoxon rank-sum test to
evaluate whether the differences in IGD values between a reference
algorithm and other algorithms are statistically significant. A low
p-value (≤ 0.05) indicates a significant difference, while a high
p-value suggests no significant difference. To convey both significance
and performance direction, we assign a `+' if an algorithm performs
significantly better than the reference, a `--' if it performs worse,
and `=' if no significant difference is observed.
\textcolor{blue}{\hyperref[_Ref215610742]{Table 5}} reports the Wilcoxon p-values and
corresponding signs for the test problems, using DPCME-RF3 as the
reference algorithm for 72-bar and 120-bar truss problems. For the
72-bar truss case, all algorithms perform significantly worse than
DPCME-RF3, except NSGAIIARSBX and MTCMO, which show no significant
difference. In the 120-bar truss case, most algorithms exhibit inferior
performance, with MTCMO showing no significant difference and
NSGAIIARSBX achieving significantly better results. For the tanker
problem, all algorithms perform worse than the reference algorithm
DPCME-RF1, except NSGAIIARSBX, which shows no significant difference.
One should keep in mind that these tests were performed on the final
populations, i.e., at generation 500. If they were performed at earlier
stages of the optimization, e.g., at generation 50 or 100, the proposed
DPCME and its repair variants would significantly outperform all other
algorithms. Thus, the proposed algorithm not only achieves better or
comparable results to the best performing algorithms in the world, but
is much better at early stages, which is important in real-life
scenarios with limited resources.

\begin{center}
\protect\phantomsection\label{_Ref215610686}{}Table 4. Friedman test
results based on final populations.
\end{center}

\makebox[\linewidth][c]{\includegraphics[width=6.0in,height=6.0in]{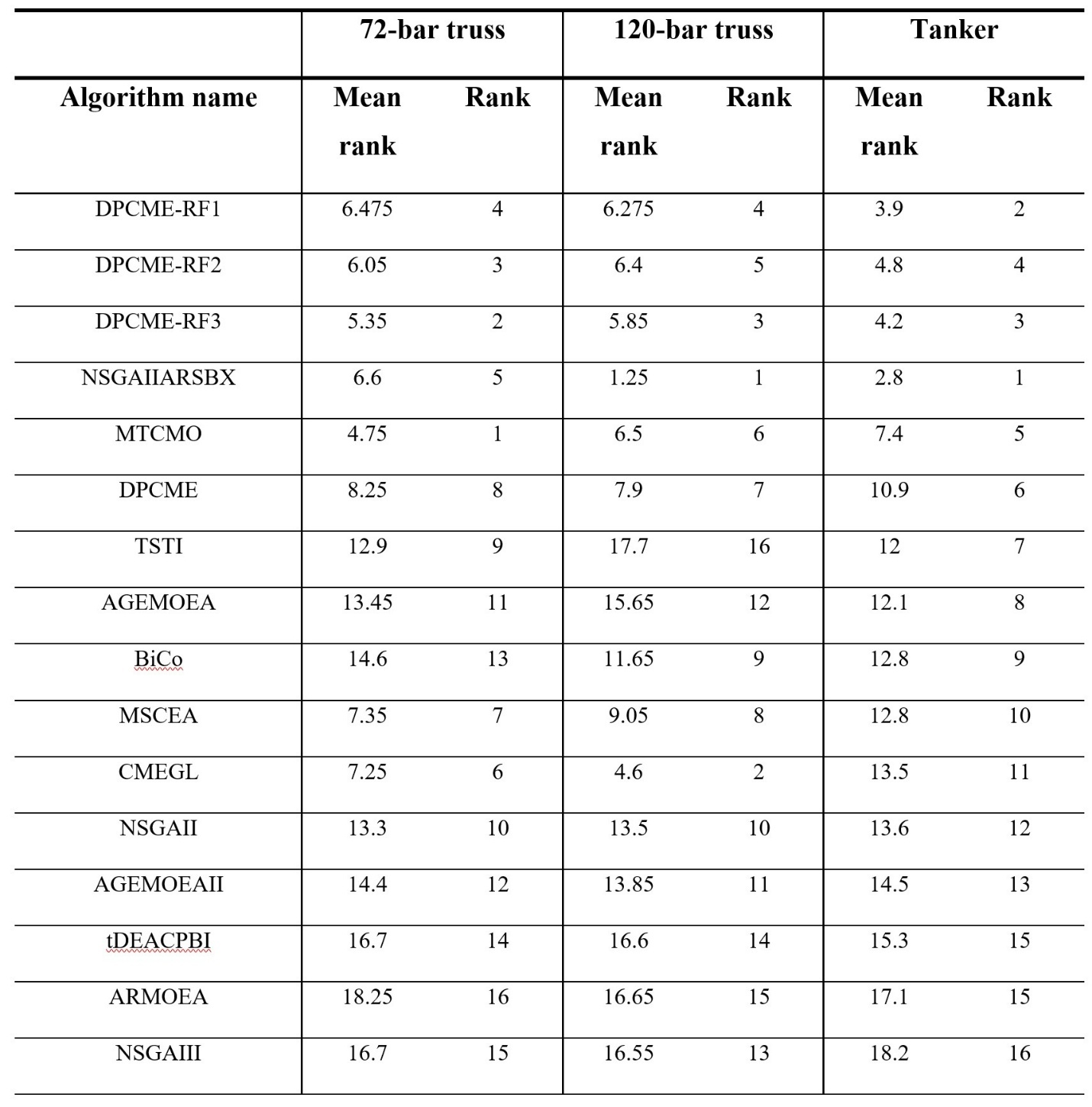}}

\begin{center}
\protect\phantomsection\label{_Ref215610742}{}Table 5. Wilcoxon p-values
calculated with 5\% significance level on final populations. Reference
algorithm for the truss problems is DPCME-RF3 and for the tanker is
DPCME-RF1.
\end{center}

\makebox[\linewidth][c]{\includegraphics[width=6.5in,height=5.35486in]{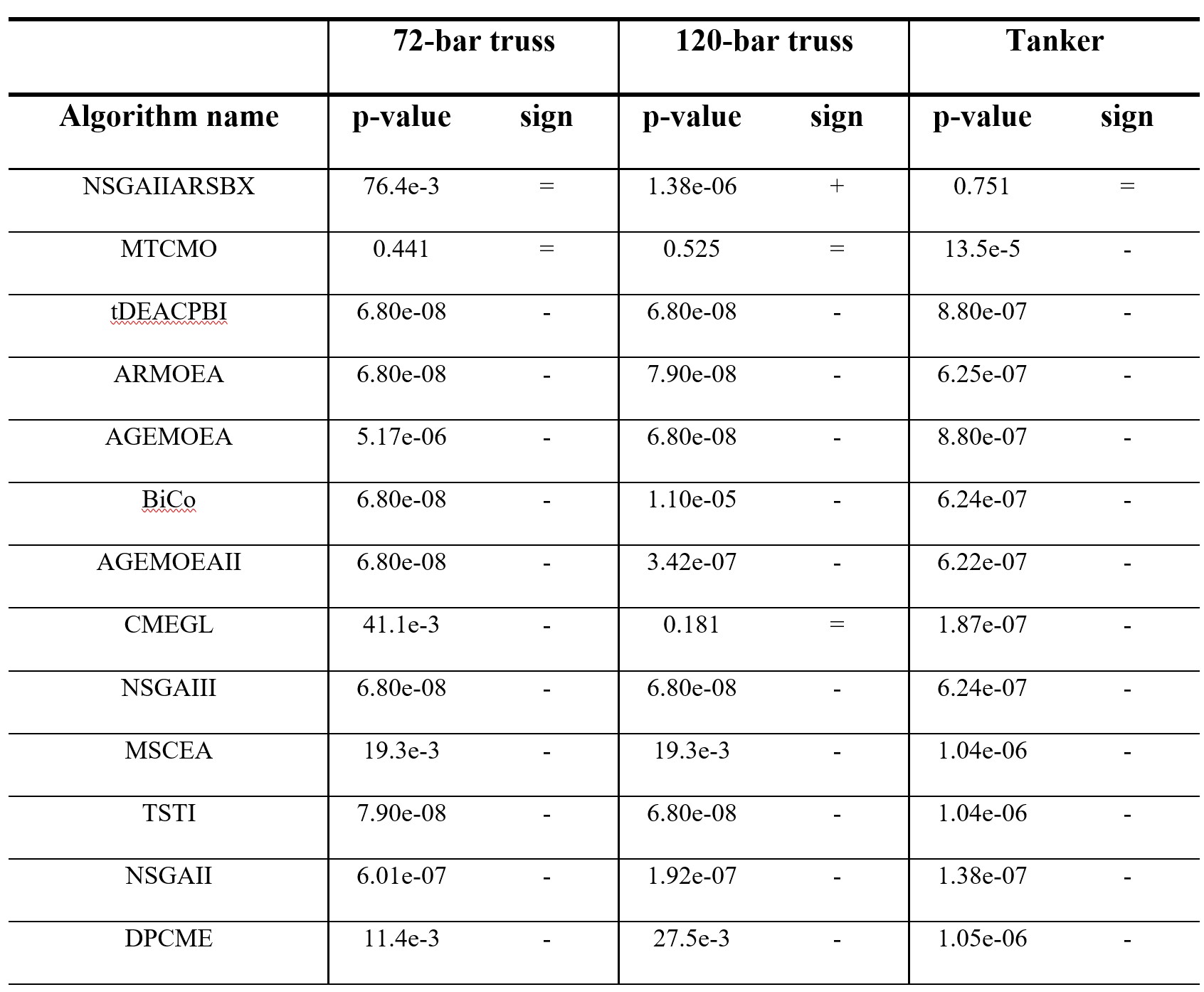}}

\section{Conclusion}\label{conclusion}

A new constrained multi-objective optimization algorithm is proposed,
called the Dual-Population Constrained Multi-objective Evolutionary
(DPCME) algorithm. By employing two populations and enabling knowledge
transfer between them, DPCME effectively reduces the likelihood of
premature convergence and getting trapped in local minima. Furthermore,
a few new strategies for repair constraint handling are proposed, where
feasible solutions are created utilizing information available in the
current populations, avoiding user-defined heuristic on how to create a
feasible design. The repair CHT can be used on a small portion of the
population, significantly enhancing algorithm's convergence speed and
overall performance.

We tested the baseline DPCME and a few variants with different repair
strategies on three complex engineering cases: the 72-bar truss, 120-bar
truss, and tanker problems, all of which involve between 144 and 376
nonlinear structural failure constraints. For comparison, twelve
best-performing state-of-the-art algorithms from the PlatEMO package
were also evaluated on the same cases, following initial comparison with
43 algorithms. The results demonstrate that DPCME, both with and without
repair, consistently ranks among the best-performing algorithms. The
proposed repair approaches further improve performance by exploiting a
larger pool of solutions during the fully infeasible phase (RI) and by
prioritizing proximity in objective space rather than crowding distance.
In addition, allowing infeasible solutions to be used for repair in the
later RF stages can lead to slight performance improvements. Considering
practical time constraints in real-world engineering optimization, we
further evaluated performance at initial optimization stage (20\% of
total generations): DPCME with repair significantly outperformed the
other algorithms. A direct comparison between DPCME with and without the
repair indicates that the repair CHT substantially improves performance
across all test cases. Among the various repair strategies, comparing
the results of different repair strategies showed that RF3 usually
emerges as the most effective across all test cases.

The results indicate that NSGAIIARSBX and MTCMO achieve strong
performance on complex optimization problems. As a next step, selected
mechanisms and techniques from these algorithms could be incorporated
into DPCME to further enhance its effectiveness. In addition, the repair
CHT occasionally generates some infeasible solutions; this behavior
warrants further investigation and presents an opportunity for future
refinements.

\section{Acknowledgments}\label{acknowledgments}

The funding for this research has been kindly provided by Seaspan
Vancouver Shipyards and Natural Sciences and Engineering Research
Council of Canada (NSERC), grants RGPIN-2025-04421 and
DGDND-2025-04421. This research was supported in part through the
computational resources and services provided by Advanced Research
Computing at The University of British Columbia. The authors also thank Robert Allan Ltd. personnel for their constructive comments and support.

\section{References}\label{references}

\begin{enumerate}
\def\labelenumi{[\arabic{enumi}]}
\item \hypertarget{ref:1}{}
  M. Dehghani, Z. Montazeri, E. Trojovská, P. Trojovský, Coati
  Optimization Algorithm: A new bio-inspired metaheuristic algorithm for
  solving optimization problems, Knowl. Based Syst. 259 (2023) 110011.
  https://doi.org/10.1016/j.knosys.2022.110011.
\item \hypertarget{ref:2}{}
  M. Azizi, S. Talatahari, A.H. Gandomi, Fire Hawk Optimizer: A novel
  metaheuristic algorithm, Artif. Intell. Rev. 56 (2023) 287--363.
  https://doi.org/10.1007/s10462-022-10173-w.
\item \hypertarget{ref:3}{}
  B. Abdollahzadeh, F.S. Gharehchopogh, N. Khodadadi, S. Mirjalili,
  Mountain Gazelle Optimizer: A new nature-inspired metaheuristic
  algorithm for global optimization problems, Adv. Eng. Softw. 174
  (2022) 103282. https://doi.org/10.1016/j.advengsoft.2022.103282.
\item \hypertarget{ref:4}{}
  J.S. Pan, L.G. Zhang, R.B. Wang, V. Snášel, S.C. Chu, Gannet
  Optimization Algorithm: A new metaheuristic algorithm for solving
  engineering optimization problems, Math. Comput. Simul. 202 (2022)
  343--373. https://doi.org/10.1016/j.matcom.2022.06.007.
\item \hypertarget{ref:5}{}
  L. Abualigah, D. Yousri, M.A. Elaziz, A.A. Ewees, M.A.A. Al-qaness,
  A.H. Gandomi, Aquila Optimizer: A novel meta-heuristic optimization
  algorithm, Comput. Ind. Eng. 157 (2021) 107250.
  https://doi.org/10.1016/j.cie.2021.107250.
\item \hypertarget{ref:6}{}
  F.A. Hashim, E.H. Houssein, K. Hussain, M.S. Mabrouk, W. Al-Atabany,
  Honey Badger Algorithm: New metaheuristic algorithm for solving
  optimization problems, Math. Comput. Simul. 192 (2022) 84--110.
  https://doi.org/10.1016/j.matcom.2021.08.013.
\item \hypertarget{ref:7}{}
  Kaveh, A. Dadras, A novel meta-heuristic optimization algorithm:
  Thermal Exchange Optimization, Adv. Eng. Softw. 110 (2017) 69--84.
  https://doi.org/10.1016/j.advengsoft.2017.03.014.
\item \hypertarget{ref:8}{}
  S. Talatahari, M. Azizi, M. Tolouei, B. Talatahari, P. Sareh, Crystal
  Structure Algorithm (CryStAl): A metaheuristic optimization method,
  IEEE Access 9 (2021) 71244--71261.
  https://doi.org/10.1109/ACCESS.2021.3079161.
\item \hypertarget{ref:9}{}
  S. Yacoubi, G. Manita, A. Chhabra, O. Korbaa, S. Mirjalili, A
  multi-objective Chaos Game Optimization algorithm based on
  decomposition and random learning mechanisms for numerical
  optimization, Appl. Soft Comput. 144 (2023) 110525.
  https://doi.org/10.1016/j.asoc.2023.110525.
\item \hypertarget{ref:10}{}
  M. Azizi, Atomic Orbital Search: A novel metaheuristic algorithm,
  Appl. Math. Model. 93 (2021) 657--683.
  https://doi.org/10.1016/j.apm.2020.12.021.
\item \hypertarget{ref:11}{}
  F.A. Hashim, E.H. Houssein, M.S. Mabrouk, W. Al-Atabany, S. Mirjalili,
  Henry Gas Solubility Optimization: A novel physics-based algorithm,
  Future Gener. Comput. Syst. 101 (2019) 646--667.
  https://doi.org/10.1016/j.future.2019.07.015.
\item \hypertarget{ref:12}{}
  T.S.L.V. Ayyarao et al., War Strategy Optimization Algorithm: A new
  effective metaheuristic algorithm for global optimization, IEEE Access
  10 (2022) 25073--25105. https://doi.org/10.1109/ACCESS.2022.3153493.
\item \hypertarget{ref:13}{}
  L. Abualigah, A. Diabat, S. Mirjalili, M.A. Elaziz, A.H. Gandomi, The
  Arithmetic Optimization Algorithm, Comput. Methods Appl. Mech. Eng.
  376 (2021) 113609. https://doi.org/10.1016/j.cma.2020.113609.
\item \hypertarget{ref:14}{}
  A.I.J. Forrester, A.J. Keane, Recent advances in surrogate-based
  optimization, Prog. Aerosp. Sci. 45 (2009) 50--79.
\item \hypertarget{ref:15}{}
  D.R. Jones, M. Schonlau, W.J. Welch, Efficient global optimization of
  expensive black-box functions, J. Glob. Optim. 13 (1998) 455--492.
\item \hypertarget{ref:16}{}
  R.G. Regis, C.A. Shoemaker, A stochastic radial basis function method
  for the global optimization of expensive functions, INFORMS J. Comput.
  19 (2007) 497--509.
\item \hypertarget{ref:17}{}
  H.-M. Gutmann, A radial basis function method for global optimization,
  J. Glob. Optim. 19 (2001) 201--227.
\item \hypertarget{ref:18}{}
  J. Knowles, ParEGO: A hybrid algorithm with on-line landscape
  approximation for expensive multiobjective optimization problems, IEEE
  Trans. Evol. Comput. 10 (2006) 50--66.
\item \hypertarget{ref:19}{}
  Y. Jin, A comprehensive survey of fitness approximation in
  evolutionary computation, Soft Comput. 9 (2005) 3--12.
  https://doi.org/10.1007/s00500-003-0329-1.
\item \hypertarget{ref:20}{}
  W. Li, R. Mai, Z. Wang, Y. Qiu, B. Xu, Z. Hao, Z. Fan,
  Surrogate-assisted push and pull search for expensive constrained
  multi-objective optimization problems, Swarm Evol. Comput. 91 (2024)
  101728. https://doi.org/10.1016/j.swevo.2024.101728.
\item \hypertarget{ref:21}{}
  D.E. Goldberg, Genetic Algorithms in Search, Optimization and Machine
  Learning, Addison-Wesley, 1989.
\item \hypertarget{ref:22}{}
  J.H. Holland, Adaptation in Natural and Artificial Systems, University
  of Michigan Press, Ann Arbor, 1975.
\item \hypertarget{ref:23}{}
  R. Storn, K. Price, Differential evolution -- A simple and efficient
  heuristic for global optimization over continuous spaces, J. Glob.
  Optim. 11 (1997) 341--359. https://doi.org/10.1023/A:1008202821328.
\item \hypertarget{ref:24}{}
  Rechenberg, Evolutionsstrategie: Optimierung technischer Systeme nach
  Prinzipien der biologischen Evolution, Frommann-Holzboog, Stuttgart,
  1973.
\item \hypertarget{ref:25}{}
  N. Hansen, A. Ostermeier, Completely derandomized self-adaptation in
  evolution strategies, Evol. Comput. 9 (2001) 159--195.
  https://doi.org/10.1162/106365601750190398.
\item \hypertarget{ref:26}{}
  Z.-H. Zhan, J. Li, J. Cao, J. Zhang, H.S.-H. Chung, Y.-H. Shi,
  Multiple populations for multiple objectives: A coevolutionary
  technique for solving multiobjective optimization problems, IEEE
  Trans. Cybern. 43 (2013) 445--463.
  https://doi.org/10.1109/TSMCB.2012.2209115.
\item \hypertarget{ref:27}{}
  J. Xiao, W. Li, B. Liu, P. Ni, A novel multi-population coevolution
  immune optimization algorithm, Soft Comput. 20 (2016) 3657--3671.
  https://doi.org/10.1007/s00500-015-1724-3.
\item \hypertarget{ref:28}{}
  J. Shi, M. Gong, W. Ma, L. Jiao, A multipopulation coevolutionary
  strategy for multiobjective immune algorithm, Sci. World J. (2014)
  Article ID 539128. https://doi.org/10.1155/2014/539128.
\item \hypertarget{ref:29}{}
  K. Qiao et al., Evolutionary constrained multiobjective optimization:
  Scalable high-dimensional constraint benchmarks and algorithm, IEEE
  Trans. Evol. Comput. (2023).
  https://doi.org/10.1109/TEVC.2023.3281666.
\item \hypertarget{ref:30}{}
  K. Qiao, J. Liang, Z. Liu, K. Yu, C. Yue, B. Qu, Evolutionary
  multitasking with global and local auxiliary tasks for constrained
  multi-objective optimization, IEEE/CAA J. Autom. Sin. 10 (2023)
  1951--1964. https://doi.org/10.1109/JAS.2023.123336.
\item \hypertarget{ref:31}{}
  P. Koch, S. Bagheri, W. Konen, C. Foussette, P. Krause, T. Bäck, A new
  repair method for constrained optimization, in: Proc. GECCO 2015, ACM,
  New York, 2015, pp. 273--280. https://doi.org/10.1145/2739480.2754658.
\item \hypertarget{ref:32}{}
  H. Ozbasaran, A kinematic stability repair algorithm for planar truss
  topology via geometric decomposition, Comput. Struct. 244 (2021)
  106428. https://doi.org/10.1016/j.compstruc.2020.106428.
\item \hypertarget{ref:33}{}
  L. Zhu, J. Lin, Y.Y. Li, Z.J. Wang, A decomposition-based
  multi-objective genetic programming hyper-heuristic approach for the
  multi-skill resource constrained project scheduling problem, Knowl.
  Based Syst. 225 (2021) 107099.
  https://doi.org/10.1016/j.knosys.2021.107099.
\item \hypertarget{ref:34}{}
  H. Tang, J. Lee, Adaptive initialization LSHADE algorithm enhanced
  with gradient-based repair for real-world constrained optimization,
  Knowl. Based Syst. 246 (2022) 108696.
  https://doi.org/10.1016/j.knosys.2022.108696.
\item \hypertarget{ref:35}{}
  Todoroki, R.T. Haftka, Stacking sequence optimization by a genetic
  algorithm with a new recessive gene like repair strategy, Compos. Part
  B Eng. 29 (1998) 277--285.
  https://doi.org/10.1016/S1359-8368(97)00030-9.
\item \hypertarget{ref:36}{}
  F. Samanipour, J. Jelovica, Adaptive repair method for constraint
  handling in multi-objective genetic algorithm based on relationship
  between constraints and variables, Appl. Soft Comput. 90 (2020)
  106143. https://doi.org/10.1016/j.asoc.2020.106143.
\item \hypertarget{ref:37}{}
  Y. Cai, J. Jelovica, Neural network-enabled discovery of mapping
  between variables and constraints for autonomous repair-based
  constraint handling in multi-objective structural optimization, Knowl.
  Based Syst. 280 (2023) 111032.
  https://doi.org/10.1016/j.knosys.2023.111032.
\item \hypertarget{ref:38}{}
  J. Jelovica, Y. Cai, Improved multi-objective structural optimization
  with adaptive repair-based constraint handling, Eng. Optim. 56 (2024)
  118--137. https://doi.org/10.1080/0305215X.2022.2147518.
\item \hypertarget{ref:39}{}
  Y. Tian, R. Cheng, X. Zhang, Y. Jin, PlatEMO: A MATLAB platform for
  evolutionary multi-objective optimization, IEEE Comput. Intell. Mag.
  12 (2017) 73--87.
\item \hypertarget{ref:40}{}
  K. Deb, A. Pratap, S. Agarwal, T. Meyarivan, A fast and elitist
  multiobjective genetic algorithm: NSGA-II, IEEE Trans. Evol. Comput. 6
  (2002) 182--197. https://doi.org/10.1109/4235.996017.
\item \hypertarget{ref:41}{}
  M. Asafuddoula, T. Ray, R. Sarker, K. Alam, An adaptive constraint
  handling approach embedded MOEA/D, in: Proc. IEEE CEC 2012, IEEE,
  2012, pp. 1--8. https://doi.org/10.1109/CEC.2012.6252868.
\item \hypertarget{ref:42}{}
  E. Zitzler, M. Laumanns, L. Thiele, SPEA2: Improving the strength
  pareto evolutionary algorithm, 2001.
  https://api.semanticscholar.org/CorpusID:16584254 (accessed 2 January
  2026).
\item \hypertarget{ref:43}{}
  J. Hernández-Díaz, A. Quesada, E. Urbina, E. Sánchez, R. Ruiz, E.
  Alba, An efficient non-dominated sorting method for evolutionary
  algorithms, Comput. Oper. Res. 35 (2008) 3701--3714.
  https://doi.org/10.1016/j.cor.2007.12.010.
\item \hypertarget{ref:44}{}
  P. Bader, A. Zitzler, Non-dominated sorting methods for
  multi-objective optimization: Review and numerical comparison, J. Ind.
  Manag. Optim. 16 (2020) 1617--1643.
  https://doi.org/10.3934/jimo.2020009.
\item \hypertarget{ref:45}{}
  Y. Tian, H. Wang, X. Zhang, Y. Jin, Effectiveness and efficiency of
  non-dominated sorting for evolutionary multi- and many-objective
  optimization, Complex Intell. Syst. 3 (2017) 247--263.
  https://doi.org/10.1007/s40747-017-0057-5.
\item \hypertarget{ref:46}{}
  L. Lamberti, An efficient simulated annealing algorithm for design
  optimization of truss structures, Comput. Struct. 86 (2008)
  1936--1953. https://doi.org/10.1016/j.compstruc.2008.02.004.
\item \hypertarget{ref:47}{}
  S.O. Degertekin, G. Yalcin Bayar, L. Lamberti, Parameter-free Jaya
  algorithm for truss sizing-layout optimization under natural frequency
  constraints, Comput. Struct. 245 (2021) 106461.
  https://doi.org/10.1016/j.compstruc.2020.106461.
\item \hypertarget{ref:48}{}
  S.H.S. Taheri, S. Jalili, Enhanced biogeography-based optimization: A
  new method for size and shape optimization of truss structures with
  natural frequency constraints, Lat. Am. J. Solids Struct. 13 (2016)
  1406--1430. https://doi.org/10.1590/1679-78252208.
\item \hypertarget{ref:49}{}
  A. Kaveh, V.R. Mahdavi, Multi-objective colliding bodies optimization
  algorithm for design of trusses, J. Comput. Des. Eng. 6 (2019) 49--59.
  https://doi.org/10.1016/j.jcde.2018.04.001.
\item \hypertarget{ref:50}{}
  A. Klanac, S. Ehlers, J. Jelovica, Optimization of crashworthy marine
  structures, Mar. Struct. 22 (2009) 670--690.
  https://doi.org/10.1016/j.marstruc.2009.06.002.
\item \hypertarget{ref:51}{}
  J. Jelovica, A. Klanac, Multi-objective optimization of ship
  structures: Using guided search vs. conventional concurrent
  optimization, in: Proc. of Marstruct 2009, 2nd International
  Conference on Marine Structures Analysis and Design of Marine, 2009,
  pp. 447--456.
\item \hypertarget{ref:52}{}
  A. Klanac, J. Jelovica, Vectorization in the Structural Optimization
  of a Fast Ferry, Brodogradnja. 58 (2007) 11--17.
  https://hrcak.srce.hr/11580
\item \hypertarget{ref:53}{}
  A. Klanac, J. Jelovica, M. Niemeläinen, S. Damagallo, H. Remes, J.
  Romanoff, Structural omni-optimization of a tanker, Proc. 7th Int.
  Conf. Comput. Appl. Inf. Technol. Marit. Ind. (COMPIT) (2008).
\item \hypertarget{ref:54}{}
  A. Klanac, J. Jelovica, Vectorization and constraint grouping to
  enhance optimization of marine structures, Mar. Struct. 22 (2009)
  225--245. https://doi.org/10.1016/j.marstruc.2008.07.001
\item \hypertarget{ref:55}{}
  O.F. Hughes, Ship Structural Design: A Rationally-Based,
  Computer-Aided Optimization Approach, 2nd ed., Society of Naval
  Architects and Marine Engineers, Jersey City, NJ, 1988.
\item \hypertarget{ref:56}{}
  British Steel, Bulb Flats,
  https://britishsteel.co.uk/media/41nb1mjk/bulb-flats-brochure.pdf
  (accessed 2 January 2026).
\item \hypertarget{ref:57}{}
  O.F. Hughes, B. Ghosh, Y. Chen, Improved prediction of simultaneous
  local and overall buckling of stiffened panels, Thin-Walled Struct. 42
  (2004) 827--856. https://doi.org/10.1016/j.tws.2004.01.003.
\item \hypertarget{ref:58}{}
  H. Naar, P. Varsta, P. Kujala, A theory of coupled beams for strength
  assessment of passenger ships, Mar. Struct. 17 (2004) 590--611.
  https://doi.org/10.1016/j.marstruc.2005.03.004.
\item \hypertarget{ref:59}{}
  O.F. Hughes, F. Mistree, V. Zanic, A practical method for the rational
  design of ship structures, J. Ship Res. 24 (1980) 101--113.
  https://doi.org/10.5957/jsr.1980.24.2.101.
\item \hypertarget{ref:60}{}
  C.A. Coello Coello, M. Reyes Sierra, A study of the parallelization of
  a coevolutionary multi-objective evolutionary algorithm, in: Proc.
  PPSN VIII, Springer, 2004, pp. 688--697.
  https://doi.org/10.1007/978-3-540-24694-7\_71.
\item \hypertarget{ref:61}{}
  R. Li, Z. Shari, M.Z.A. Ab Kadir, A review on multi-objective
  optimization of building performance -- Insights from bibliometric
  analysis, Heliyon 11 (2025) e42480.
  https://doi.org/10.1016/j.heliyon.2025.e42480.
\item \hypertarget{ref:62}{}
  K. Qiao et al., Dynamic auxiliary task-based evolutionary multitasking
  for constrained multi-objective optimization, IEEE Trans. Evol.
  Comput. 27 (2023) 642--656. https://doi.org/10.1109/TEVC.2022.3153107.
\item \hypertarget{ref:63}{}
  L. Pan, W. Xu, L. Li, C. He, R. Cheng, Adaptive simulated binary
  crossover for rotated multi-objective optimization, Swarm Evol.
  Comput. 60 (2021) 100759. https://doi.org/10.1016/j.swevo.2020.100759.
\end{enumerate}

\end{document}